\documentclass[11pt]{article}

\usepackage[preprint]{acl}

\usepackage{times}
\usepackage{latexsym}

\usepackage[T1]{fontenc}

\usepackage[utf8]{inputenc}

\usepackage{microtype}

\usepackage{inconsolata}

\usepackage{graphicx}
\usepackage{graphics}
\usepackage[utf8]{inputenc} 
\usepackage[T1]{fontenc}    
\usepackage{url}            
\usepackage{booktabs}       
\usepackage{amsfonts}       
\usepackage{nicefrac}       
\usepackage{microtype}      
\usepackage{xcolor}         
\usepackage[utf8]{inputenc}
\usepackage{graphicx}
\usepackage{amsmath}
\usepackage{amssymb}
\usepackage{mathtools}
\usepackage{amsthm}
\usepackage{arydshln}
\usepackage{multirow}
\usepackage{wrapfig, lipsum, booktabs}
\usepackage{algorithm}
\usepackage{enumitem}
\usepackage{paralist, tabularx}
\usepackage{balance}
\usepackage[noend]{algpseudocode}
\usepackage{pgfplots}
\usetikzlibrary{pgfplots.groupplots}
\pgfplotsset{compat=1.3}
\usepackage{tikz}
\usetikzlibrary{patterns}
\usepackage{pgf-pie}
\usepackage{adjustbox}
\usepackage{colortbl}
\usepackage{xspace}
\usepackage{titling}
\usepackage{xcolor}
\usepackage{enumitem}
\usepackage[table]{xcolor}
\usepackage[most]{tcolorbox}

\definecolor{demphcolor}{RGB}{144, 144, 144}
\definecolor{mygray}{gray}{0.4}
\definecolor{lightgray}{rgb}{0.9, 0.9, 0.9}

\definecolor{dt}{HTML}{ADCAD8}
\definecolor{dt2}{HTML}{cddfe7}

\usepackage{pifont}
\usepackage{float}
\definecolor{my_green}{RGB}{51,102,0}
\definecolor{my_red}{RGB}{204, 0, 0}

\DeclareMathAlphabet{\mathsfit}{\encodingdefault}{\sfdefault}{m}{sl}
\SetMathAlphabet{\mathsfit}{bold}{\encodingdefault}{\sfdefault}{bx}{n}

\definecolor{oorange}{RGB}{252,218,227}
\definecolor{yyellow}{RGB}{255,237,203}
\definecolor{ppurple}{RGB}{208,205,226}
\definecolor{ggreen}{RGB}{195,222,176}
\definecolor{ggrey}{RGB}{230,230,230}
\definecolor{rred}{RGB}{247,187,187}
\definecolor{wwhite}{RGB}{255,255,255}
\definecolor{pairaware}{RGB}{67, 130, 180}      
\definecolor{jointtrain}{RGB}{230, 145, 56}      
\definecolor{adaptiverl}{RGB}{76, 153, 76}       

\newcommand{\modelname}{MMEmb-R1\xspace}

\definecolor{newcontent}{RGB}{0,120,0}      
\definecolor{movedcontent}{RGB}{0,80,200}   

\newcommand{\add}[1]{{\color{black}#1}}
\newcommand{\revise}[1]{{\color{black}#1}}
\title{\modelname: Reasoning-Enhanced Multimodal Embedding with Pair-Aware Selection and Adaptive Control
}

\newcommand*\samethanks[1][\value{footnote}]{\footnotemark[#1]}

\author{Yuchi Wang$^{12}$\thanks{~~Work done during an internship at ByteDance.}, ~Dingkang Yang$^{13}$, ~Haiyang Yu$^{2}$, ~Weikang Bian$^{1}$, ~Jiefeng Long$^{2}$,\\ \textbf{Xiao Liang$^{2}$,~Chao Feng$^{2}$\thanks{~~Chao Feng and Hongsheng Li are corresponding authors.}, ~Hongsheng Li$^{1}$\samethanks} \\
 $^{1}$MMLab, The Chinese University of Hong Kong 
 \\
    {$^{2}$ByteDance Inc. $~~~$  
    $^{3}$Fudan University
    }\\
     \texttt{wangyuchi369@gmail.com~chaofeng.zz@bytedance.com~hsli@ee.cuhk.edu.hk} \\
   ~\\
   ~\\
}

\makeatletter
\AtBeginDocument{
\let\@oldmaketitle\@maketitle
\renewcommand{\@maketitle}{\@oldmaketitle
  \vspace{-28pt}
  \includegraphics[width=1\linewidth]{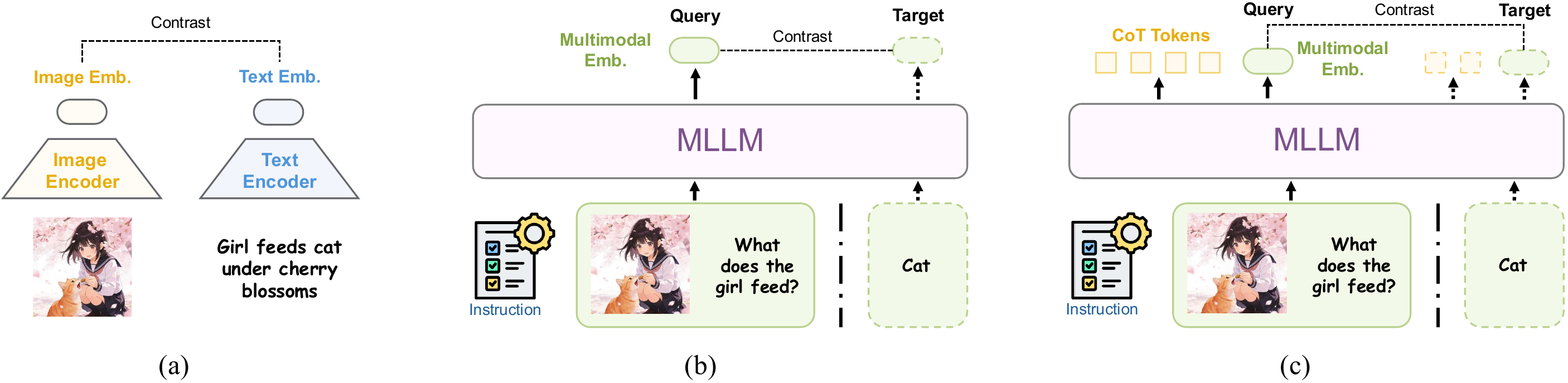}
  \vspace{-18pt}
  \captionof{figure}{The evolution of multimodal embedding. (a) Early approaches employ modality-specific encoders to project different modalities into a shared semantic space. (b) MLLM-based methods process multimodal inputs with task instructions and are trained using semantically related query--target pairs. (c) Recent reasoning-enhanced embedding methods introduce chain-of-thought (CoT) reasoning prior to generating multimodal embeddings.
}
  \label{fig:intro}
  \vspace{32pt}
 }
}
\makeatother

\begin{document}
\maketitle

\begin{abstract}

MLLMs have been successfully applied to multimodal embedding tasks, yet their generative reasoning capabilities remain underutilized. 
Directly incorporating chain-of-thought reasoning into embedding learning introduces two fundamental challenges. 
First, structural misalignment between instance-level reasoning and pairwise contrastive supervision may lead to shortcut behavior, where the model merely learns the superficial format of reasoning. 
Second, reasoning is not universally beneficial for embedding tasks. Enforcing reasoning for all inputs may introduce unnecessary computation and latency, and can even obscure salient semantic signals for simple cases.
To address these issues, we propose \modelname, an adaptive reasoning-based multimodal embedding framework. 
\revise{We formulate reasoning as a latent variable and introduce pair-aware reasoning selection
to identify reasoning paths beneficial for query--target alignment. }
Furthermore, we adopt reinforcement learning to selectively invoke reasoning only when necessary. 
Experiments on the MMEB-V2 benchmark demonstrate that our model achieves a score of 71.2 with only 4B parameters, establishing a new state-of-the-art while significantly reducing reasoning overhead and inference latency.

\end{abstract}

\section{Introduction}~\label{sec:intro}

Multimodal embedding models aim to project heterogeneous inputs, such as text, images, and interleaved image-text content, into a unified semantic space. 
They serve as a fundamental infrastructure for a wide range of applications, including recommendation systems~\cite{lin2025sailembeddingtechnicalreportomnimodal,zhang2025notellm2multimodallargerepresentation}, cross-modal retrieval~\cite{copali,uniir}, and retrieval-augmented generation~\cite{yu2025visragvisionbasedretrievalaugmentedgeneration,ren2025videoragretrievalaugmentedgenerationextreme}. 
Early work, exemplified by CLIP~\cite{radford2021learningtransferablevisualmodels}, leverages large-scale image-text pairs to align different modalities within a shared semantic space (Fig.~\ref{fig:intro}(a)).
More recently, multimodal large language models (MLLMs) have revolutionized this field~\cite{mmebv2,zhang2025gmeimprovinguniversalmultimodal,vlm2vec} by providing rich world knowledge, compositional understanding, and strong instruction-following capabilities (Fig.~\ref{fig:intro}(b)).

However, the current utilization of MLLMs in embedding models remains limited. 
Most existing approaches treat MLLMs primarily as static feature extractors, without fundamentally departing from the conventional paradigm. 
In contrast, the success of LLMs~\cite{brown2020languagemodelsfewshotlearners,yang2025qwen3technicalreport} and MLLMs~\cite{comanici2025gemini25pushingfrontier,openai2024gpt4ocard} largely stems from their generative capability: next-token prediction and the generative paradigm have substantially enhanced abstraction, reasoning, and structured understanding, giving rise to emergent abilities~\cite{wei2022emergentabilitieslargelanguage}. 
The embedding community has only marginally benefited from these strengths. 
This raises a fundamental question: 
\emph{Can generative reasoning be effectively integrated into embedding learning, and if so, what is the appropriate formulation?}

\begin{figure}
    \centering
    \includegraphics[width=1\linewidth]{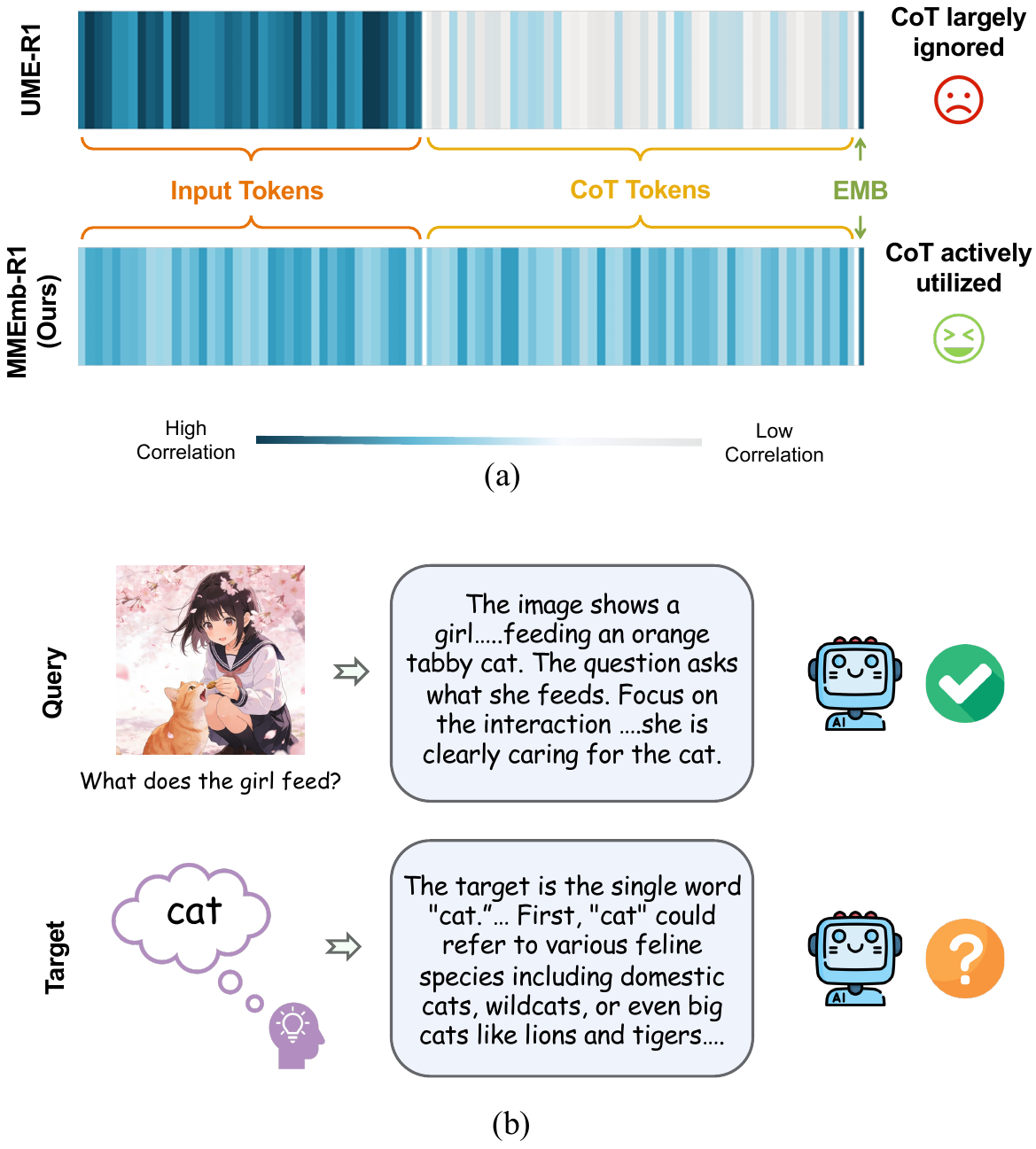}
    \caption{Two challenges of reasoning in embedding. (a) Shortcut behavior: UME-R1's embedding token largely ignores CoT tokens, while \modelname{} actively utilizes them. (b) Overthinking: reasoning helps the complex query (top) but introduces irrelevant noise for the simple target ``cat'' (bottom).}
    \label{fig:challenge}
\end{figure}

By reexamining the paradigms of embedding and reasoning, as well as prior related works, we identify two key challenges. 
\textbf{(1) Structural misalignment between reasoning and representation learning may induce shortcut behavior.} Embedding models are trained under pairwise contrastive supervision, whereas reasoning is generated at the instance level.
Existing pioneering reasoning-driven embedding models~\cite{ume,tte}, as illustrated in Fig.~\ref{fig:intro}(c), typically require the model to learn or incorporate a single teacher-provided chain-of-thought (CoT)~\cite{wei2023chainofthoughtpromptingelicitsreasoning} separately for the query and the target before generating the embedding. 
In this setup, reasoning quality is largely decoupled from the paired objective that governs contrastive representation learning. 
As shown in Fig.~\ref{fig:challenge}(a), embedding tokens in prior models such as UME-R1~\cite{ume} attend heavily to the original input but minimally to CoT tokens, suggesting that reasoning is often treated as a deterministic procedural prefix rather than a latent variable subject to selection. 
Consequently, the model exhibits shortcut behavior: it mimics the surface format of reasoning without establishing a meaningful dependency between reasoning and the learned representation. 
\textbf{(2) Reasoning is not universally necessary for embedding tasks.} 
For simple or concise inputs, enforced autoregressive reasoning may induce ``overthinking'', introducing unnecessary computation and latency~\cite{kim2025letmultimodalembedderslearn}. 
Moreover, excessive reasoning can obscure salient semantic signals and may even degrade performance by confusing the model, as shown in Fig.~\ref{fig:challenge}(b).

To harvest the ability of the generative paradigm and address above challenges, 
we propose \modelname, an adaptive \textbf{R}easoning-based \textbf{M}ulti\textbf{M}odal 
\textbf{Emb}edding framework. 
Instead of deterministically generating a single reasoning trajectory, we formulate 
the reasoning path as a latent variable and introduce \textbf{a pair-aware reasoning 
selection mechanism tailored to contrastive embedding}. 
Specifically, we employ multiple heterogeneous worker MLLMs to generate diverse 
reasoning candidates, simulating a rich prior distribution over the latent reasoning 
space and mitigating single-teacher bias. 
\revise{We then design a pair-aware evaluator 
to score each reasoning path: by comparing the matching confidence with and without 
the rationale, we isolate its marginal contribution to query-target alignment, which subsequently guides model training.}
Furthermore, we develop \textbf{an adaptive reasoning mechanism} that explicitly models 
the utility of reasoning and mitigates unnecessary overthinking. 
We quantify the reasoning benefit 
by computing the similarity gap between reasoning-enhanced and direct embeddings. 
This continuous utility signal serves as a reward in reinforcement learning with 
GRPO~\cite{Guo_2025}, enabling the model to learn a policy that selectively invokes reasoning only 
when it provides substantial benefit. 
By integrating pair-aware selection with adaptive reasoning control, our framework 
achieves a principled balance between effectiveness and efficiency.

Extensive experiments on the MMEB-V2 benchmark~\citep{mmebv2} demonstrate the effectiveness of our approach.
\modelname{} achieves state-of-the-art performance across both small-size and medium-size settings, attaining 68.3 overall with a Qwen3-VL-2B backbone and 71.2 with Qwen3-VL-4B, surpassing strong baselines such as Embed-RL~\cite{embedrl} (66.8) and RzenEmbed-v1~\cite{jian2025rzenembedcomprehensivemultimodalretrieval} (68.9) while using fewer parameters.
\revise{The proposed adaptive mechanism  increases inference throughput by $3.1\times$ compared to UME-R1~\cite{ume}, with improvement in retrieval accuracy.}
We hope this work offers a fresh perspective on reasoning-aware representation learning and opens new avenues for integrating generative paradigms into multimodal embedding.
\section{Related Works}

\subsection{Multimodal Embedding Models}

Multimodal embedding aims to learn compact, semantically meaningful representations for heterogeneous data.
CLIP~\cite{radford2021learningtransferablevisualmodels} established the dual-tower contrastive paradigm, training separate encoders via large-scale image-text alignment.
Subsequent studies extended this paradigm to additional modalities like AudioCLIP~\cite{guzhov2021audioclipextendingclipimage} and CLIP4Clip~\cite{luo2021clip4clipempiricalstudyclip}. 
Other works further improve the contrastive learning paradigm by introducing novel training objectives or pre-training strategies, such as BLIP~\cite{li2022blipbootstrappinglanguageimagepretraining} and SigLIP~\cite{zhai2023sigmoidlosslanguageimage}.
With the rise of MLLMs, the community has shifted toward MLLM-based embedding frameworks.
Early representative works include VLM2Vec~\cite{vlm2vec}, GME~\cite{zhang2025gmeimprovinguniversalmultimodal}, and ColPali~\cite{faysse2025colpaliefficientdocumentretrieval}. 
Building on this foundation, recent efforts have explored expanding modality coverage~\cite{mmebv2,jian2025rzenembedcomprehensivemultimodalretrieval,tzachor2026vidvec,liu2025rematch}, scaling data quality~\cite{li2026qwen3vlembeddingqwen3vlrerankerunifiedframework,zhou2024megapairsmassivedatasynthesis,gu2025unimev2mllmasajudgeuniversalmultimodal}, and designing specialized architectures or training strategies~\cite{chen2025mocamodalityawarecontinualpretraining,qin2025unimocounifiedmodalitycompletion,gu2026mucomultiturncontrastivelearning,li2026magic}.
More recently, several studies have explored incorporating generative reasoning into embedding learning.
UME-R1~\cite{ume} applies supervised fine-tuning to endow embedding models with reasoning capability;
TTE~\cite{tte} investigates diverse combinations of reasoners and embedders;
and our concurrent work Embed-RL~\cite{embedrl} optimizes the reasoner to generate evidential chains of thought.
While these pioneering efforts demonstrate the potential of reasoning for embedding, they largely overlook the structural misalignment between instance-level reasoning and pair-level contrastive supervision, which motivates the design of \modelname{}.

\subsection{Large Reasoning Models}
Recent advances have shown that LLMs and MLLMs benefit substantially from enhanced reasoning capabilities~\cite{Guo_2025, comanici2025gemini25pushingfrontier}, as exemplified by OpenAI o1~\cite{openai2024openaio1card} and QwQ~\cite{qwq32b}.
Early methods adopt chain-of-thought prompting~\cite{kojima2023largelanguagemodelszeroshot,wang2023selfconsistencyimproveschainthought,xu2025llavacotletvisionlanguage,shao2024visualcotadvancingmultimodal} to elicit step-by-step rationales.
Inspired by GRPO in DeepSeek-R1~\cite{Guo_2025}, a growing body of work applies reinforcement learning to optimize reasoning trajectories across diverse domains, including visual understanding~\cite{feng2025videor1reinforcingvideoreasoning,shen2025vlmr1stablegeneralizabler1style}, text-to-image generation~\cite{jiang2025t2ir1reinforcingimagegeneration}, mathematical reasoning~\cite{lu2024mathvistaevaluatingmathematicalreasoning,zhang2024mavismathematicalvisualinstruction}, and domain-specific applications such as finance~\cite{liu2026finr1largelanguagemodel} and medicine~\cite{lai2025medr1reinforcementlearninggeneralizable}.
However, multimodal embedding and representation learning, an important subfield of multimodal learning, has yet to benefit from this paradigm much, a gap our work aims to bridge.
\section{Methodology}\label{sec:method}

\begin{figure*}
    \centering
    \includegraphics[width=0.96\linewidth]{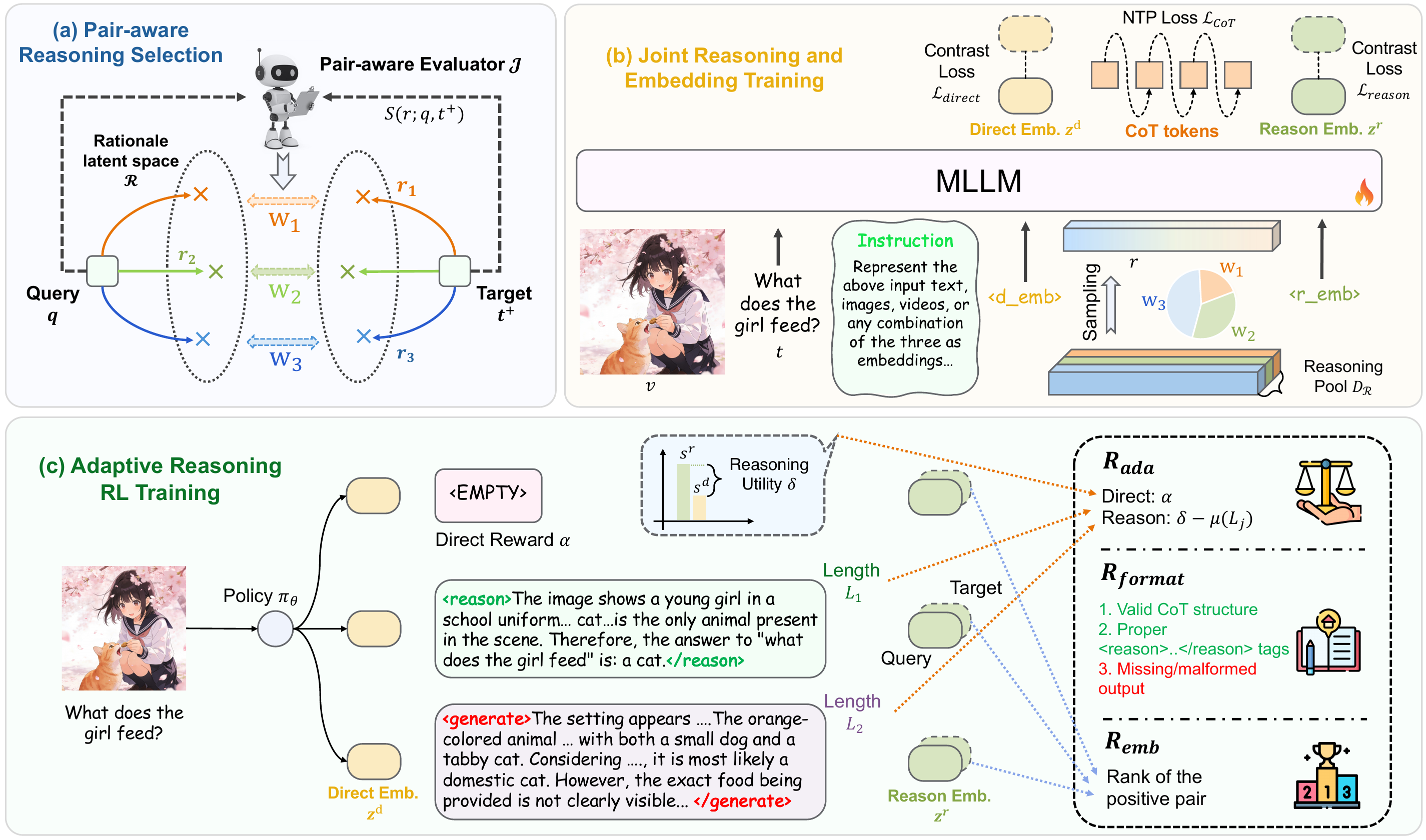}
    \caption{Overview of the \modelname{} framework. \textcolor{pairaware}{Upper Left:} Pair-aware reasoning selection---multiple heterogeneous workers generate diverse rationale candidates for the query and target, and an evaluator scores each candidate to produce selection weights $w_1, w_2, w_3$. \textcolor{jointtrain}{Upper right:} Joint reasoning and embedding training---the MLLM is trained with a direct embedding path ($\mathcal{L}_{\text{direct}}$), a reasoning-enhanced embedding path ($\mathcal{L}_{\text{reason}}$), and a next-token prediction objective over CoT tokens ($\mathcal{L}_{\text{CoT}}$). \textcolor{adaptiverl}{Lower:} Adaptive reasoning via GRPO---the policy $\pi_\theta$ decides whether to invoke reasoning or emit \texttt{<EMPTY>} for each query, guided by three reward signals: adaptive reward $R_{\text{ada}}$, format reward $R_{\text{format}}$, and embedding reward $R_{\text{emb}}$.}
    \label{fig:method}
\end{figure*}

We present our \modelname framework, illustrated in Fig.~\ref{fig:method}, which consists of three stages: (1) constructing a pair-aware reasoning pool via diverse candidate generation and \revise{posterior} selection (\textsection~\ref{sec:pair_aware}); (2) jointly training the model for reasoning generation and contrastive embedding (\textsection~\ref{sec:joint_training}); and (3) adaptive reasoning control via utility-aware reinforcement learning (\textsection~\ref{sec:adaptive_rl}). We begin with preliminaries and an architectural overview in \textsection~\ref{sec:prelim}.

\subsection{Preliminaries and Framework Overview}\label{sec:prelim}

\paragraph{Preliminaries of Multimodal Embedding.}
Given a multimodal input $x = \{t, v\}$ consisting of text and visual (image or video) content, an embedding model $\mathcal{E}$ maps it to a $d$-dimensional representation $\mathbf{z} = \mathcal{E}(x) \in \mathbb{R}^d$.
Training follows the contrastive paradigm: given a batch of $N$ query--target pairs $\{(q_k, t_k^+)\}_{k=1}^N$, we compute embeddings $\mathbf{z}_{q_k} = \mathcal{E}(q_k)$ and $\mathbf{z}_{t_k} = \mathcal{E}(t_k^+)$. The objective pulls positive pairs closer while pushing in-batch negatives apart by optimizing the InfoNCE loss:
\[
\mathcal{L}_{\text{con}} = 
-\frac{1}{N} \sum_{k=1}^N
\log
\frac{\exp\bigl(\mathrm{sim}(\mathbf{z}_{q_k}, \mathbf{z}_{t_k})/\tau\bigr)}
{\sum_{j=1}^N \exp\bigl(\mathrm{sim}(\mathbf{z}_{q_k}, \mathbf{z}_{t_j})/\tau\bigr)}
\]
where $\tau$ is sampling temperature and $\mathrm{sim}(\cdot,\cdot)$ denotes cosine similarity.

\paragraph{Architecture Overview.}
\modelname is built upon a multimodal large language model (MLLM).
Visual inputs are first processed by a vision transformer (ViT)~\cite{Dosovitskiy2020AnII} and projected into the language token space via a visual adapter, enabling unified sequence modeling across modalities.
The model operates in two modes.
In \emph{direct mode}, the embedding is extracted from the hidden state of the final input special token: $\mathbf{z}^{\text{d}} = \mathcal{E}(x)$.
In \emph{reasoning mode}, the model first generates a reasoning path $r$ conditioned on the input, and the embedding is derived from the final token after the reasoning trajectory: $\mathbf{z}^{\text{r}} = \mathcal{E}(x \oplus r)$, where $\oplus$ denotes sequence concatenation.

\paragraph{Reasoning as a Latent Variable.}
A central departure of \modelname from prior work is the treatment of the reasoning path $r$ as a \emph{latent variable} rather than a deterministic output of a fixed teacher.
Formally, we posit a latent reasoning space $\mathcal{R}$ with a prior distribution $\mathcal{P}(R)$, from which reasoning candidates are sampled: $r \sim \mathcal{P}(R)$.
The reasoning-enhanced embedding can then be written as a marginalization over this latent space:
$
\mathbf{z}^{\text{r}} = \mathbb{E}_{r \sim \mathcal{P}(R)}\bigl[\mathcal{E}(x \oplus r)\bigr].
$
In practice, direct marginalization is intractable.
Our framework addresses this by: \textbf{(1)} simulating $\mathcal{P}(R)$ through diverse multi-worker generation, \textbf{(2)} introducing a pair-aware scoring function $S(r; q, t^+)$ to perform structured posterior selection aligned with the contrastive objective, and \textbf{(3)} learning an adaptive policy that decides whether to sample from $\mathcal{P}(R)$ at all.
We detail each component in the following sections.

\subsection{Pair-Aware Reasoning Selection for Contrastive Embedding}\label{sec:pair_aware}

As established in \textsection~\ref{sec:prelim}, we model reasoning as a latent variable $r \sim \mathcal{P}(R)$ whose quality should be assessed under the joint query--target context.

\subsubsection{Diverse Prior Simulation via Multi-Worker Generation}\label{sec:candidate_gen}

To approximate a rich prior $\mathcal{P}(R)$ and reduce single-teacher bias, we employ $K$ heterogeneous worker MLLMs $\{M_k\}_{k=1}^K$ spanning complementary capabilities:
\textbf{(1)} Instruct-based models (e.g., Qwen2-VL-Instruct~\cite{wang2024qwen2vlenhancingvisionlanguagemodels}): produce concise, structured analyses of core semantics and retrieval-relevant keypoints.
\textbf{(2)} Thinking models (e.g., GLM-4.1V-Thinking~\cite{vteam2026glm45vglm41vthinkingversatilemultimodal}): generate exploratory, long-form reasoning chains that capture deeper analytical perspectives, though potentially with greater verbosity.
\textbf{(3)} High-capacity proprietary models (e.g., Gemini 2.5 Pro~\cite{comanici2025gemini25pushingfrontier}): provide broad world knowledge and rich contextual coverage. As shown in Fig.~\ref{fig:method}(a), for each input $x$ (either a query $q$ or target $t^+$), each worker independently produces a candidate rationale $r_k = M_k(x)$, $k = 1, \dots, K$.
Note that generation is still performed \emph{single-sided} in this stage to avoid information leakage.
The resulting candidates $\mathcal{R}_x = \{r_1, r_2, \dots, r_K\}$ collectively form empirical samples from the latent reasoning prior $\mathcal{P}(R)$. Detailed prompt and implementations can be found in Appendix.~\ref{sec:app_worker_prompt}.

\subsubsection{\revise{Posterior Selection}}\label{sec:scoring}

Given samples from the prior, we perform \emph{posterior selection}: identifying which reasoning paths are most useful for the pair $(q_i, t_i^+)$.
Specifically, we employ an evaluator model $\mathcal{J}$ prompted to judge whether the query and target match, and extract the logit of the affirmative token \texttt{[YES]} as a confidence score.
\revise{Inspired by causal intervention~\cite{pearl2009causality}, we isolate the contribution of reasoning by comparing the matching confidence computed without and with a rationale candidate: $c_0 = \mathrm{Conf}_{\mathcal{J}}(q_i, t_i^+)$ and $c_r = \mathrm{Conf}_{\mathcal{J}}(q_i, t_i^+, r)$.
The \emph{reasoning gain} is: $\label{eq:delta}
\Delta_r = c_r - c_0. $, measuring how much rationale $r$ improves recognizing query-target correspondence beyond raw input.
Positive $\Delta_r$ indicates useful semantic bridging rather than mere rephrasing. This method can also serve as baseline calibration: subtracting the sample-wise ``counterfactual baseline'' $c_0$ yields a more normalized and robust scoring signal across different samples.}
We retain candidates with $\Delta_r > \epsilon$, forming $\mathcal{R}_i^+ = \{r \in \mathcal{R}_i \mid \Delta_r > \epsilon\}$, and normalize gains via softmax:
\[
w_{r} = \frac{\exp(\Delta_r / \gamma)}{\sum_{r' \in \mathcal{R}_i^+} \exp(\Delta_{r'} / \gamma)}
\]
where $\gamma$ is a temperature controlling the sharpness of the selection distribution.
This produces a weighted reasoning pool:
$\mathcal{D}_R = \bigl\{(q_i, t_i^+, r_{i,j}, w_{i,j})\bigr\}_{i,j}$
where higher-gain reasoning paths contribute more strongly to subsequent training. More details can be found in Appendix~\ref{sec:app_evaluator} and Appendix~\ref{app:gain_dist}.

\subsection{Joint Reasoning and Embedding Training}\label{sec:joint_training}

With the curated reasoning pool $\mathcal{D}_R$ representing the selected posterior over latent reasoning, we fine-tune the MLLM to acquire: (1) contrastive alignment for embedding matching, and (2) coherent chain-of-thought generation internalizing the reasoning distribution.
This is achieved through a multi-objective training scheme with two complementary embedding paths (Fig.~\ref{fig:method}(b)).

\paragraph{Reasoning-Enhanced Embedding Path.}
For each training pair $(q_i, t_i^+)$, we sample a reasoning path $r_{i,j}$ from $\mathcal{D}_R$ according to its posterior weight $w_{i,j}$.
This path is optimized with the contrastive loss:
$
\mathcal{L}_{\text{reason}} = \mathcal{L}_{\text{con}}\bigl(\mathbf{z}^{\text{r}}_{q}, \mathbf{z}^{\text{r}}_{t}\bigr),
$
where $\mathcal{L}_{\text{con}}$ follows the InfoNCE formulation defined above.
To explicitly cultivate reasoning generation ability within the backbone, we additionally apply a next-token prediction loss over the chain-of-thought tokens:
\[
\mathcal{L}_{\text{CoT}} = -\sum_{l=1}^{|r|} \log p_\theta(r_l \mid x, r_{<l}),
\]
which trains the model to internalize the reasoning trajectories in $\mathcal{D}_R$ as generative knowledge.

\paragraph{Direct Embedding Path.}
To preserve embedding quality without reasoning overhead, we include a direct path encoding raw inputs as $\mathbf{z}^{\text{d}} = \mathcal{E}(x)$, optimized with:
$
\mathcal{L}_{\text{direct}} = \mathcal{L}_{\text{con}}\bigl(\mathbf{z}^{\text{d}}_{q}, \mathbf{z}^{\text{d}}_{t}\bigr).
$

\paragraph{Overall Objective.}
The complete training objective combines all three components:
\[
\mathcal{L} = \mathcal{L}_{\text{reason}} + \lambda_{CoT} \mathcal{L}_{\text{CoT}} + \lambda_{direct} \mathcal{L}_{\text{direct}}.
\]
where $L_{CoT}$ and $L_{direct}$ are hyperparameters.

\subsection{Adaptive Reasoning Control via Utility-Aware Optimization}\label{sec:adaptive_rl}

While the joint training stage equips the model with reasoning capability, not all inputs benefit from explicit reasoning as discussed in \textsection~\ref{sec:intro}.
We therefore introduce a reinforcement learning stage that trains the model to \emph{selectively invoke reasoning only when it provides measurable benefit}.

\subsubsection{Reasoning Utility Estimation}\label{sec:utility}

We estimate reasoning utility from the embedding geometry learned after the joint training stage. 
For each query $q_i$ in the reinforcement learning dataset, we compute its similarity with the corresponding target using both normalized direct embeddings and reasoning-enhanced embeddings produced by the jointly trained model, yielding $s_i^{\mathrm{d}}$ and $s_i^{\mathrm{r}}$, respectively. 
We then define the \emph{reasoning utility gap} as $
\delta_i = s_i^{\mathrm{r}} - s_i^{\mathrm{d}}$.
This continuous signal quantifies the marginal benefit of reasoning for each instance: $\delta_i > 0$ indicates that reasoning improves retrieval quality, whereas $\delta_i \leq 0$ suggests that direct embedding is sufficient or even preferable. 
Importantly, we treat $\delta_i$ as a continuous intrinsic signal rather than a binary supervision label, enabling more fine-grained and stable policy learning.

\begin{table*}[t]
\centering
\renewcommand{\arraystretch}{1.2}
\resizebox{\textwidth}{!}{
\begin{tabular}{ll|ccccc|ccccc|ccccc|c}
\toprule[1.5pt]
\multirow{2}{*}{\textbf{Model}} 
& \multirow{2}{*}{\textbf{Backbone}} 
& \multicolumn{5}{c|}{\textbf{Image}} 
& \multicolumn{5}{c|}{\textbf{Video}} 
& \multicolumn{5}{c|}{\textbf{VisDoc}}
& \multirow{2}{*}{\textbf{All}}\\
\cmidrule(lr){3-7} \cmidrule(lr){8-12} \cmidrule(lr){13-17}
& &\textbf{CLS} & \textbf{QA} & \textbf{RET} & \textbf{GD} & \textbf{Overall} 
& \textbf{CLS} & \textbf{QA} & \textbf{RET} & \textbf{MRET} & \textbf{Overall} 
& \textbf{VDRv1} & \textbf{VDRv2} & \textbf{VR} & \textbf{OOD} & \textbf{Overall} \\
\midrule

\rowcolor{dt!50}
\multicolumn{18}{c}{\textbf{Small-size Models}} \\

GME & Qwen2-VL-2B & 
54.4 & 29.9 & 66.9 & 55.5 & 51.9 & 
34.9 & 42.0 & 25.6 & 32.4 & 33.9 & 
86.1 & 54.0 & 82.5 & 43.1 & 72.7 & 
54.1 \\

ColPali-V1.3 & PaliGemma-3B & 
40.3 & 11.5 & 48.1 & 40.3 & 34.9 & 
26.7 & 37.8 & 21.6 & 25.5 & 28.2 & 
83.6 & 52.0 & 81.1 & 43.1 & 71.0 & 
44.4 \\

VLM2Vec & Qwen2-VL-2B & 
58.7 & 49.3 & 65.0 & 72.9 & 59.7 & 
33.4 & 30.5 & 20.6 & 33.0 & 29.0 & 
49.8 & 13.5 & 51.8 & 33.5 & 41.6 & 
47.0 \\

VLM2Vec-V2 & Qwen2-VL-2B & 
62.9 & 56.3 & 69.5 & 77.3 & 64.9 &
39.3 & 34.3 & 28.8 & 38.5 & 34.9 & 
75.5 & 44.9 & 79.4 & 39.4 & 65.4 & 
58.0 \\

UME-R1 & Qwen2-VL-2B & 
64.8 & 62.8 & 67.6 & 77.2 & 66.6 &
44.3 & 51.2 & 32.9 & 39.7 & 42.2 & 
72.4 & 46.2 & 79.2 & 37.2 & 63.9 & 
60.1 \\

$\text{TTE}_s^{\dagger}$ &Qwen3-VL-2B &
67.9 & 66.6 & 70.2 & 84.1 & \underline{70.1} &
47.3 & 49.1 & 34.4 & 33.2 & 32.1 &
77.5 & 53.2 & 83.2 & 41.1 & 68.8 &
63.1 \\

RzenEmbed-v1 & Qwen2-VL-2B&
65.3 & 61.7 & 73.8 & 77.9 & 68.5 & 
45.6 & 47.5 & 38.3 & 36.7 & 42.6 &
87.0 & 57.6 & 85.4 & 43.3 & \textbf{74.4} &
64.4 \\

Embed-RL & Qwen3-VL-2B & 
62.8 & 67.9 & 68.6 & 90.4 & 69.2 & 
57.0 & 55.9 & 45.1 & 49.4 & \underline{52.1} & 
79.9 & 52.0 & 84.6 & 65.7 & \underline{74.1} & 
\underline{66.8} \\ 


\modelname (Ours) & Qwen2-VL-2B & 
64.5 & 68.1 & 70.0 & 88.9 & 70.0 & 
56.3 & 52.8 & 47.6 & 42.5 & 50.6 & 
74.1 & 56.0 & 78.9 & 48.3 & 68.0 & 
65.0 \\ 

\modelname (Ours) & Qwen3-VL-2B & 
63.5 & 73.7 & 70.2 & 89.8 & \textbf{71.5} & 
59.8 & 60.3 & 50.6 & 49.2 & \textbf{55.6} & 
82.0 & 55.7 & 80.7 & 56.7 & 73.2 & 
\textbf{68.3} \\ 

\midrule

\rowcolor{dt!50}
\multicolumn{18}{c}{\textbf{Medium-size Models}} \\

\midrule

GME & Qwen2-VL-7B & 
57.7 & 34.7 & 71.2 & 59.3 & 56.0 & 
37.4 & 50.4 & 28.4 & 38.2 & 38.6 & 
89.4 & 55.6 & 85.0 & 44.4 & 75.2 & 
57.8 \\

LamRA & Qwen2-VL-7B & 
59.2 & 26.5 & 70.0 & 62.7 & 54.1 & 
39.3 & 42.6 & 24.3 & 34.6 & 35.2 & 
22.0 & 11.5 & 37.4 & 21.0 & 23.9 & 
40.4 \\

LamRA & Qwen2.5-VL-7B & 
51.7 & 34.1 & 66.9 & 56.7 & 52.4 & 
32.9 & 42.6 & 23.2 & 37.6 & 33.7 & 
56.3 & 33.3 & 58.2 & 40.1 & 50.2 & 
47.4 \\

VLM2Vec & Qwen2-VL-7B & 
62.7 & 56.9 & 69.4 & 82.2 & 65.5 & 
39.1 & 30.0 & 29.0 & 40.6 & 34.0 & 
56.9 & 9.4 & 59.1 & 38.1 & 46.4 & 
52.3 \\

CAFe & LLaVA-OV-7B & 
63.6 & 61.7 & 69.1 & 87.6 & 67.6 & 
35.8 & 58.7 & 34.4 & 39.5 & 42.4 & 
70.7 & 49.6 & 79.5 & 38.1 & 63.9 & 
60.6 \\

UME-R1 & Qwen2-VL-7B & 
67.1 & 69.2 & 71.9 & 84.9 & 71.3 &
48.6 & 60.7 & 38.2 & 39.3 & 47.5 & 
75.7 & 50.5 & 83.7 & 37.6 & 67.1 & 
64.5 \\

Embed-RL & Qwen3-VL-4B & 
63.7 & 70.5 & 71.3 & 91.4 & 70.1 & 
57.6 & 58.4 & 45.1 & 49.5 & 53.0 & 
80.2 & 53.4 & 84.9 & 67.1 & 74.7 & 
68.1 \\ 

$\text{TTE}_s^{\dagger}$ & Qwen2-VL-7B &
69.7 & 72.4 & 74.0 & 90.6 & 74.2 &
49.1 & 60.6 & 36.4 & 37.2 & 46.8 &
84.1 & 62.7 & 91.9 & 47.6 & 76.4 &
68.6 \\

RzenEmbed-v1 &  Qwen2-VL-7B &
69.8 & 68.7 & 76.8 & 85.7 & 73.6 &
52.8 & 56.2 & 41.9 & 41.8 & 48.9 &
89.5 & 60.8 & 87.9 & 44.4 & \textbf{76.8} &
68.9 \\

\modelname (Ours) & Qwen2-VL-7B & 68.1 & 71.8 & 77.7 & 87.1 & \underline{74.4} & 60.2 & 56.0 & 51.2 & 41.1 & \underline{53.3} & 79.6 & 65.3 & 84.7 & 58.0 & 74.9 &
\underline{69.7} \\ 

\modelname (Ours) & Qwen3-VL-4B & 
67.7 & 74.2 & 74.5 & 94.9 & \textbf{74.8} & 
60.7 & 61.1 & 51.6 & 50.6 & \textbf{56.6} & 
83.0 & 60.6 & 83.7 & 66.6 & \underline{76.7} & 
\textbf{71.2} \\ 

\bottomrule[1.5pt]
\end{tabular}
}
\caption{Comparison of baseline methods and \modelname{} on MMEB-V2. Given the diversity of model backbones, we aggregate results by model size. Models with 2B--3B parameters are categorized as \emph{small}, while those with 4B--7B parameters are categorized as \emph{medium}. $\dagger$ indicates that for the TTE model, we adopt the student variant to ensure a fair comparison without relying on a large external teacher model. 
Metrics are abbreviated as follows: \textbf{CLS} (classification), \textbf{QA} (question answering), \textbf{RET} (retrieval), \textbf{GD} (grounding), \textbf{MRET} (moment retrieval), \textbf{VDR} (ViDoRe), \textbf{VR} (VisRAG), and \textbf{OOD} (out-of-domain).}
\label{tab:main_result}
\end{table*}

\subsubsection{Policy Optimization with GRPO}\label{sec:grpo}

We formulate adaptive reasoning as a sequential decision-making problem~\cite{puterman1990markov,chen2023towards}. As shown in Fig.~\ref{fig:method}(c), for each query $q_i$, the model selects an action $a_i \in \{\textsc{Direct}, \textsc{Reason}\}$, indicating whether to generate the embedding directly or to invoke reasoning before embedding. 
If the \textsc{Reason} action is selected, the model first generates a rationale and then produces the embedding conditioned on it.
We design a reward function that balances retrieval improvement and computational cost:
\[
R_{\text{ada}} =
\begin{cases}
\alpha, & a_i = \textsc{Direct} \;\wedge\; (n \le N) \\
\delta_i - \mu(L_i), & a_i = \textsc{Reason}
\end{cases}
\]
where $\alpha$ is a positive constant that encourages exploration of the \textsc{Direct} action during the early stage of training (i.e., when the training step $n \le N$), mitigating the tendency to always generate rationales inherited from the supervised stage. 
For the \textsc{Reason} action, $L_i$ denotes the length of the generated reasoning, and $\mu(\cdot)$ controls the trade-off between performance gain and computational overhead. 
In particular, we penalize excessively long rationales by applying an additional coefficient $c$ when the reasoning length exceeds 512 tokens.
Following~\citet{ume}, we also incorporate an embedding reward $R_{\text{emb}}$, which evaluates embedding quality based on the ranking position of the positive target among in-batch negatives, and a format reward $R_{\text{format}}$ to ensure that the generated rationales follow the required output structure. 
To ensure symmetry, we additionally compute the reward in the reverse direction (target $\rightarrow$ query) and take the mean of the two scores. 
More details can be found in Appendix~\ref{app:adaptive}.
The overall objective is optimized using Group Relative Policy Optimization (GRPO)~\cite{shao2024deepseekmathpushinglimitsmathematical}, which maximizes the expected reward:
\[
\max_\theta \; \mathbb{E}_{a_i \sim \pi_\theta(\cdot \mid q_i)}
\bigl[
R_{\text{ada}} + R_{\text{format}} + R_{\text{emb}}
\bigr].
\] 

\section{Experiments}
\subsection{Setup}

\subsubsection{Implementation Details}

We build \modelname{} on the Qwen-VL family. 
For diverse prior simulation, we use GLM-4.1V-Thinking~\cite{vteam2026glm45vglm41vthinkingversatilemultimodal}, InternVL3-14B-Instruct~\cite{zhu2025internvl3exploringadvancedtraining}, and Doubao-Seed-1.6-Vision~\cite{seed1.6}. 
The pair-aware evaluator $\mathcal{J}$ is Qwen3-VL-32B-Instruct~\cite{yang2025qwen3technicalreport}.
For joint training, we use batch size 32 per GPU under DeepSpeed ZeRO-3, a cosine learning rate schedule with initial rate $5\times10^{-5}$, and train for 3 epochs. 
For adaptive reasoning, we use GRPO~\cite{shao2024deepseekmathpushinglimitsmathematical} with $\alpha = 0.2$ and $\beta = 0.04$. 
All experiments run on 8$\times$ H20 90GB GPUs. 
See Appendix~\ref{sec:app_impl} for details.

\subsubsection{Training Datasets and Benchmark}

Following prior work, we use MMEB-Train~\cite{mmebv2} for training. 
After data filtering and pair-aware selection (\textsection~\ref{sec:scoring}), we obtain $\sim$1.2M samples for joint embedding and reasoning training and $\sim$10K for adaptive reasoning reinforcement learning.
For evaluation, we use MMEB-V2~\cite{mmebv2}, a comprehensive benchmark covering 78 tasks across classification, VQA, retrieval, and visual grounding. 
Following the standard evaluation protocol, we report Hit@1 for image/video tasks and NDCG@5 for visual document tasks.

\subsection{Main Results}

\paragraph{Baselines.}
We compare \modelname{} against a broad set of multimodal embedding models, including classical MLLM-based models such as GME~\cite{zhang2025gmeimprovinguniversalmultimodal}, ColPali~\cite{faysse2025colpaliefficientdocumentretrieval}, VLM2Vec~\cite{vlm2vec}, and VLM2Vec-V2~\cite{mmebv2}; recent methods like LamRA~\cite{liu2024lamralargemultimodalmodel}, CAFe~\cite{yu2025cafeunifyingrepresentationgeneration}, and RzenEmbed~\cite{jian2025rzenembedcomprehensivemultimodalretrieval}; and reasoning-driven models including UME-R1~\cite{ume}, TTE~\cite{tte}, and Embed-RL~\cite{embedrl}.
All methods are evaluated on MMEB-V2~\cite{mmebv2} across Image, Video, and VisDoc modalities (Tab.~\ref{tab:main_result}).

\paragraph{Analysis.} We can see that \modelname{} achieves state-of-the-art performance in both size categories.
With a Qwen3-VL-2B backbone, \modelname{} attains 68.3 overall, surpassing the strongest baselines Embed-RL and RzenEmbed-v1 by +1.5 and +3.9 points respectively.
Scaling to Qwen3-VL-4B yields 71.2, outperforming the best medium-size baseline RzenEmbed-v1-7B with nearly half the parameters.
Notably, \modelname{} at 2B already surpasses several 7B baselines, suggesting that high-quality reasoning can partially compensate for the capacity gap.
The improvements are consistent across modality groups but particularly pronounced on Video, where \modelname{} (Qwen3-VL-2B) achieves 55.6, outperforming Embed-RL by +3.5. 
This aligns with our expectation: video understanding demands compositional reasoning over temporal dynamics, precisely the setting where pair-aware latent reasoning provides the greatest benefit.
Detailed results and results on MMEB-V1 can be found in Appendix~\ref{app:detailed_results}. We also provide qualitative analysis in Appendix~\ref{sec:qualitative}.

\subsection{Further Analysis}
\subsubsection{Further Analysis of Adaptive Reasoning}

\begin{table}[t]
\centering
\small
\tabcolsep=0.15cm
\begin{tabular}{l c c}
\toprule[1.5pt]
\textbf{Strategy} & \textbf{Throughput} & \textbf{Accuracy} \\
\midrule
\textcolor{gray}{\modelname (Direct)} &\textcolor{gray}{22.70 queries/s}& \textcolor{gray}{59.2}\\
UME-R1 & 3.21 queries/s & 60.1 \\
\modelname (Reason) & 5.79 queries/s & 63.6 \\
\modelname (Adaptive) & 10.02 queries/s& 65.0 \\
\bottomrule[1.5pt]
\end{tabular}
\caption{Inference throughput on a subset and overall accuracy on MMEB-V2.}
\label{tab:latency}
\vspace{-1em}
\end{table}

\paragraph{Inference latency comparison.}
As discussed in \textsection~\ref{sec:adaptive_rl}, our adaptive reasoning mechanism eliminates unnecessary reasoning paths, improving performance while reducing the inference latency of reasoning-driven embedding models.
\revise{To verify this, we compare inference throughput and performance under different inference strategies. We measure end-to-end latency using a batch size of 24 on 8$\times$H20 GPUs. For a fair comparison with UME-R1-2B, we use Qwen2VL-2B as the backbone. For the ``always-reasoning'' setting, we apply reinforcement learning but set the adaptive reward $R_{\text{ada}}$ to zero. Results are summarized in Tab.~\ref{tab:latency}. MMEmb-R1 Adaptive achieves 10.02 queries per second, a 1.7$\times$ speedup over the always-reason variant and 3.1$\times$ over UME-R1, while delivering the highest accuracy. The direct variant achieves the highest throughput but suffers a severe performance drop.}
\revise{The throughput gap between MMEmb-R1 Always-reason and UME-R1 reflects the efficiency of our pair-aware reasoning selection, which produces more concise rationales than UME-R1's verbose single-teacher chains. The further improvement from always-reason to adaptive confirms that the learned policy effectively skips unnecessary reasoning for simple queries, yielding a model that is both faster and more accurate.}

\begin{figure}[t]
\centering
\includegraphics[width=\columnwidth]{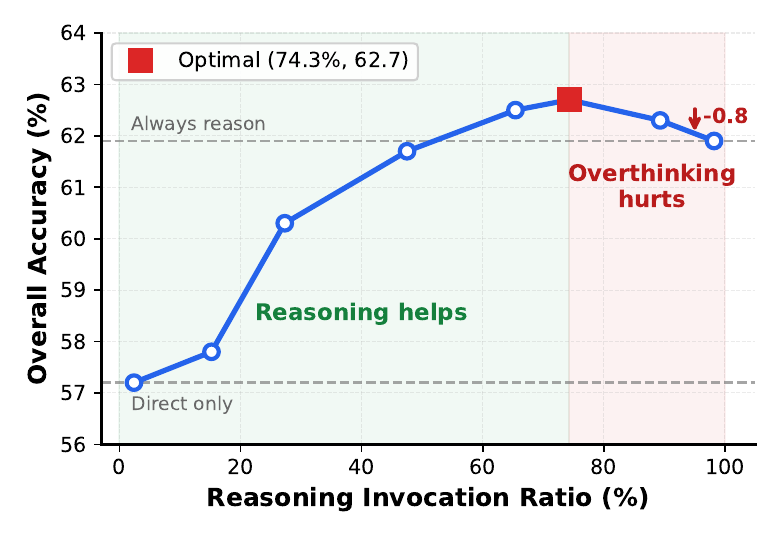}
\caption{Reasoning invocation ratio vs.\ overall accuracy. The curve peaks at 74.3\% reasoning ratio, after which accuracy declines.}
\label{fig:pareto}
\end{figure}

\paragraph{Pareto Frontier between Reasoning and Accuracy}
The coefficient $c$, which controls the trade-off between reasoning benefits and the length budget, provides an informative lens for analyzing how the adaptive policy allocates reasoning budget. 
In this experiment, we remove the 512-token limit and directly vary the cost coefficient $c$, tracing the resulting trade-off between reasoning invocation ratio and retrieval accuracy in Fig.~\ref{fig:pareto}. 
Each point corresponds to a policy trained with a different $w$ and evaluated on a subset of MMEB-V2 for efficiency.
As shown in the figure, the curve increases slowly at first and then rises steeply from around 57.2 to 62.7 at a reasoning ratio of 74.3\%. 
Beyond this point, performance declines slightly to 61.9 under near-universal reasoning, representing a 0.8-point drop that empirically confirms the overthinking phenomenon discussed in \textsection~\ref{sec:intro}. 
These results indicate that the adaptive policy learns to prioritize the most reasoning-critical instances first: the earliest queries selected for reasoning yield the highest marginal returns, while those added later contribute diminishing or even negative gains.

\begin{table}[t]
\centering

\resizebox{\columnwidth}{!}{
\begin{tabular}{l c c}
\toprule[1.5pt]
\textbf{Variant} & Score & \textbf{$\Delta$} \\
\midrule
\modelname{} (Full) & 65.0 & -- \\
\midrule
\multicolumn{3}{l}{\emph{Pair-aware Reasoning Selection}} \\
\quad Single-teacher rationale & 61.2 & $-$3.8 \\
\quad w/o pair-aware selection (uniform) & 62.8 & $-$2.2 \\
\quad w/o baseline calibration (use $c_r$ only) & 64.1 & $-$0.9 \\
\midrule
\multicolumn{3}{l}{\emph{Training Objective}} \\
\quad w/o $\mathcal{L}_{\text{reason}}$ (Direct only) & 59.2 & $-$5.8 \\
\midrule
\multicolumn{3}{l}{\emph{Adaptive Reasoning}} \\
\quad Always reason & 63.6 & $-$1.4 \\
\quad Always direct & 60.4 & $-$4.6 \\
\quad Random (50\%) & 60.6 & $-$4.4 \\
\quad Oracle & 66.2 & $+$1.2\\
\bottomrule[1.5pt]
\end{tabular}
}
\caption{Ablation study on MMEB-V2. Each row removes or modifies one component from the full \modelname{} framework.}
\label{tab:ablation}
\vspace{-1em}
\end{table}

\subsubsection{Ablation Studies}

We conduct ablation studies to validate our design choices (Tab.~\ref{tab:ablation}).
\textbf{(1)} 
Replacing the diverse multi-worker prior with a single teacher causes the largest drop, confirming that diverse candidates cover the reasoning space more effectively. 
Uniform sampling without pair-aware scoring degrades performance by 2.2 points, and removing the \revise{``counterfactual baseline'' $c_0$} causes an additional 0.9-point drop, indicating both selection and \revise{baseline calibration} filter genuinely useful rationales.
\textbf{(2)} 
Removing the reasoning path entirely (Direct only) results in a 5.8-point drop, establishing reasoning-enhanced representations as the primary driver of our framework.
\textbf{(3)} 
The always-reason variant achieves 63.6, 1.4 points lower than the full model, confirming that indiscriminate reasoning harms simple queries. 
The always-direct and random 50\% strategies perform comparably, suggesting naive stochastic selection provides no advantage over skipping reasoning, whereas the learned policy captures meaningful structure. 
The oracle strategy (selecting the better of direct vs. reasoning-enhanced embedding) provides an upper bound of 66.2, indicating our learned policy recovers most achievable gain.

\section{Conclusion}


In this paper, we present \modelname{}, a framework integrating generative reasoning into multimodal embedding learning. To address the misalignment between instance-level reasoning and pair-level contrastive supervision, we introduce a pair-aware selection mechanism. To mitigate overhead from indiscriminate reasoning, we develop a utility-aware reinforcement learning stage for selective reasoning invocation. Experiments on MMEB-V2 demonstrate state-of-the-art performance with substantially reduced inference latency. We hope this work inspires further research in reasoning-driven multimodal representation learning.

\section*{Limitations}
Our framework has several limitations that warrant future investigation.
First, the pipeline nature of our approach—offline reasoning generation, pair-aware selection, and two-stage training—prevents joint optimization of these components. An end-to-end formulation that unifies reasoning generation, selection, and adaptive invocation within a single training loop could improve overall performance.
Second, \revise{the reward design for the adaptive policy currently considers only binary decisions (whether to invoke reasoning or not), which may be suboptimal.} 
Extending the policy to control reasoning depth or granularity (e.g., brief versus detailed reasoning chains) could enable more fine-grained allocation of computational resources.
Third, reasoning-enhanced embedding inevitably incurs additional inference cost. While precomputing embeddings for the corpus side partially alleviates this in retrieval scenarios, fundamentally reducing the latency of reasoning-based models remains an open challenge.

\bibliography{custom}
\newpage
\appendix
\section{Additional Experimental Results}

\subsection{Qualitative Analysis}\label{sec:qualitative}

We conduct qualitative analyses to demonstrate the capability of \modelname{} and illustrate several design principles. 
For simplicity, we present only the main reasoning traces produced by the model. 

\begin{figure*}[t]
\centering
\includegraphics[width=0.75\linewidth]{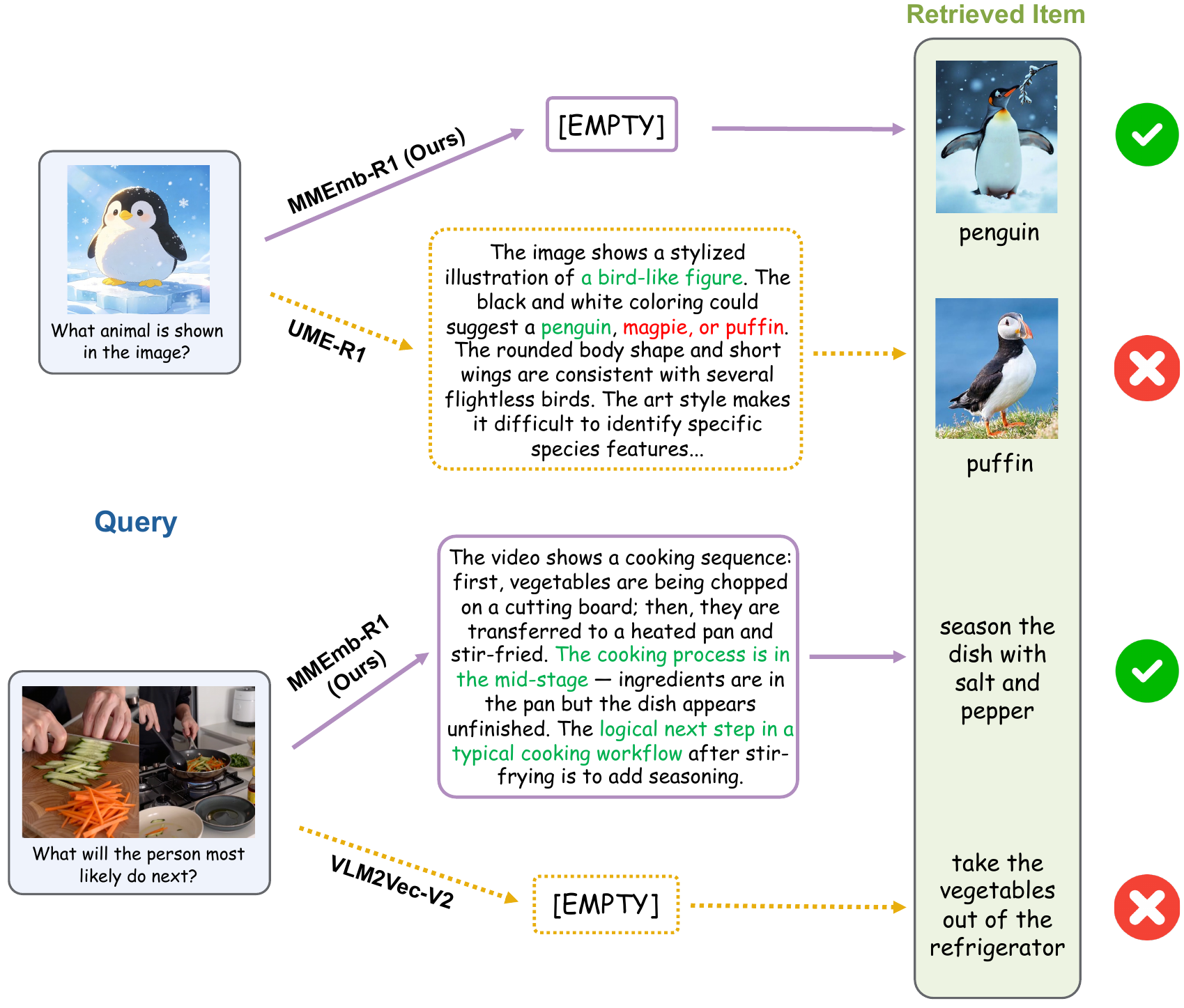}
\caption{Adaptive reasoning: \modelname{} skips reasoning for a simple visual query (top, avoiding overthinking) and invokes it for a complex temporal query (bottom), outperforming UME-R1 and VLM2Vec-V2 respectively.}
\label{fig:case_adaptive}
\end{figure*}

Fig.~\ref{fig:case_adaptive} presents two retrieval cases highlighting the advantages of adaptive reasoning in \modelname{}.
In the upper case, the query is a cartoon penguin, which is visually unambiguous. 
\modelname{} adaptively skips reasoning and correctly retrieves ``penguin,'' whereas UME-R1's enforced reasoning introduces spurious alternatives (``penguin, magpie, or puffin'') and ultimately retrieves the wrong target.
In the lower case, the query is a cooking video that requires temporal inference. 
\modelname{} invokes reasoning and correctly decomposes the cooking sequence, inferring that ``the logical next step after stir-frying is to add seasoning.'' 
In contrast, the non-reasoning baseline VLM2Vec-V2 appears to capture only the coarse semantic concept of cooking and retrieves a temporally preceding action instead. 
These examples demonstrate that \modelname{} learns when reasoning is beneficial and when it is unnecessary.

\begin{figure*}[t]
\centering
\includegraphics[width=\textwidth]{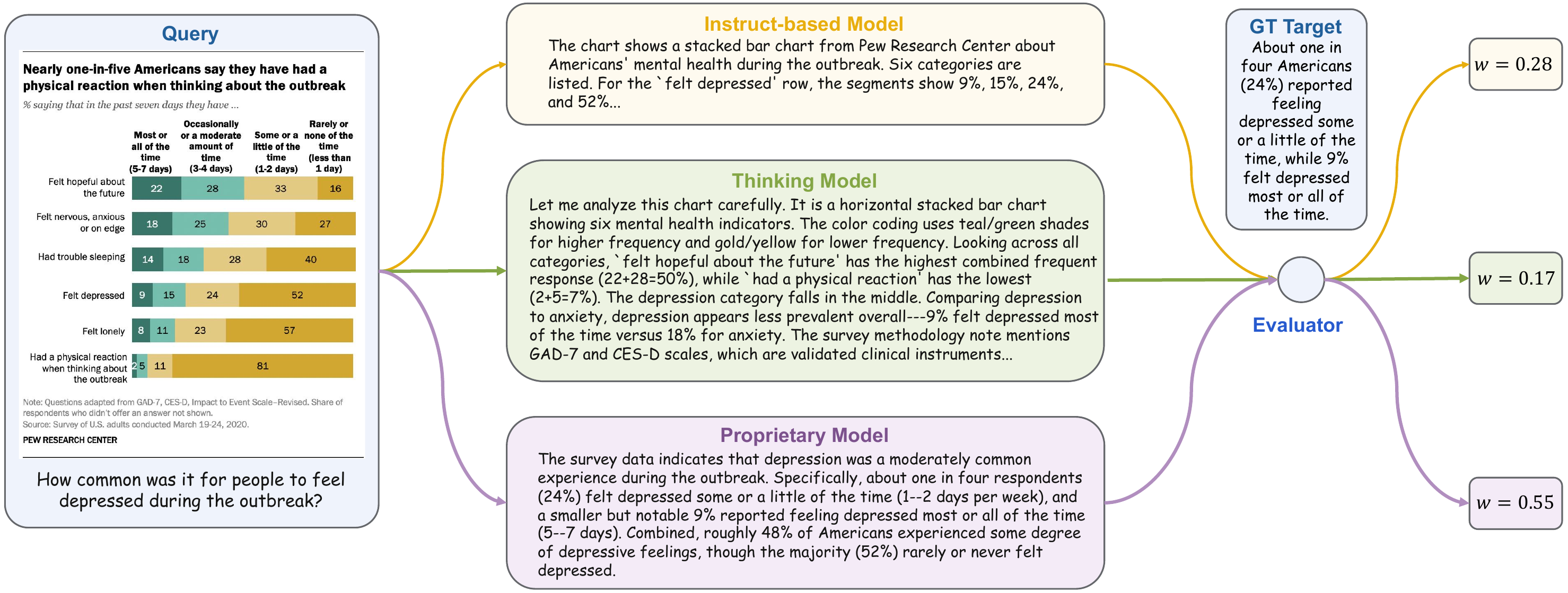}
\caption{Pair-aware reasoning selection: three heterogeneous workers produce complementary rationales for the same query. The evaluator assigns the highest weight to the Proprietary worker ($w\!=\!0.55$), which best bridges the query--target semantic gap.}
\label{fig:case_worker}
\end{figure*}

Fig.~\ref{fig:case_worker} further illustrates why diverse workers combined with pair-aware selection outperform single-source reasoning approaches such as UME-R1 and TTE. 
Given a chart query asking ``How common was it for people to feel depressed during the outbreak?'' and a ground-truth target stating that ``about one in four Americans (24\%) reported feeling depressed some or a little of the time, while 9\% felt depressed most or all of the time,'' the three workers exhibit complementary strengths and weaknesses.
The Instruct worker ($w\!=\!0.28$) correctly extracts the relevant numbers (9\%, 15\%, 24\%, 52\%) but merely lists them without interpreting which frequency band each corresponds to, leaving a gap between the raw data and the natural-language phrasing of the target. 
The Thinking worker ($w\!=\!0.17$) produces a detailed cross-category comparison—contrasting depression with anxiety and discussing clinical instruments—but this exhaustive analysis diverges from the specific query, burying the target-relevant information. 
The Proprietary worker ($w\!=\!0.55$) receives the highest weight because it directly mirrors the target semantics: it rephrases ``24\%'' as ``about one in four respondents,'' associates ``9\%'' with the ``most or all of the time'' frequency band, and provides the complementary statistic that the majority rarely felt depressed. 
These examples illustrate that the pair-aware evaluator identifies reasoning paths that effectively bridge the semantic gap between a specific query and its target, rather than favoring reasoning that is merely more elaborate or complex.

\begin{figure}[t]
\centering
\includegraphics[width=\columnwidth]{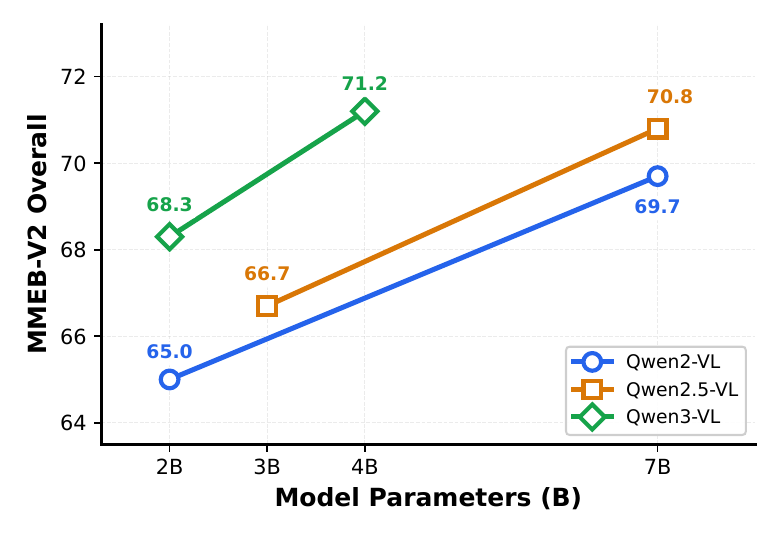}
\caption{Scaling behavior of \modelname{} across backbone families and parameter scales. Performance improves consistently within each family, and newer architectures achieve higher scores at comparable or smaller model sizes.}
\label{fig:scaling}
\end{figure}

\begin{figure}[t]
\centering
\includegraphics[width=\columnwidth]{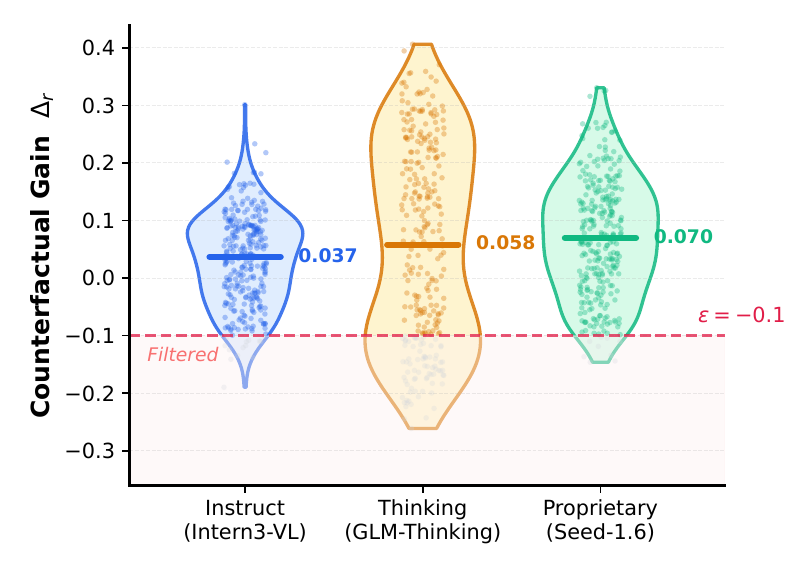}
\caption{Distribution of reasoning gains $\Delta_r$ across worker types. Median values are shown as horizontal bars. The dashed line indicates the selection threshold $\epsilon = -0.1$; candidates below are filtered out.}
\label{fig:gain_dist}
\end{figure}

\subsection{Scaling Behavior Across Backbones}\label{app:scaling}

To assess the generality and scalability of \modelname{}, we apply our framework to six backbone MLLMs spanning three model families and varying parameter scales: Qwen2-VL (2B, 7B), Qwen2.5-VL (3B, 7B), and Qwen3-VL (2B, 4B).
Fig.~\ref{fig:scaling} reports the overall MMEB-V2 performance for each configuration.

Two observations emerge.
First, \modelname{} exhibits consistent intra-family scaling: performance improves monotonically with model size across all three families, indicating that our framework effectively leverages the additional capacity of larger backbones without saturating.
Second, the gains from backbone architecture advancement are substantial and largely orthogonal to those from scaling.
Qwen3-VL-2B surpasses Qwen2-VL-7B at less than one-third of the parameters, and Qwen3-VL-4B outperforms Qwen2.5-VL-7B at roughly half the size.
This suggests that \modelname{} benefits from both stronger representations and larger capacity, and that the pair-aware reasoning selection and adaptive invocation mechanisms transfer effectively across architectures without architecture-specific tuning.

\subsection{Reasoning Gain Distribution}\label{app:gain_dist}

Fig.~\ref{fig:gain_dist} presents the distribution of ``counterfactual reasoning gains'' $\Delta_r$ across the three worker types used in our diverse prior simulation (values are rescaled for better visualization). 
We can see that no single worker dominates the distribution. 
The Proprietary worker achieves the highest median gain and exhibits the most compact, positively skewed distribution, while the Instruct worker produces a tighter but lower-centered distribution. 
In contrast, the Thinking worker shows the widest spread with a clearly bimodal shape---its upper mode reaches the highest individual $\Delta_r$ values among all workers, yet its lower mode extends well into negative territory. 
This observation supports our hypothesis that thinking models generate exploratory reasoning chains that are occasionally exceptional but frequently noisy, making them a high-variance complement to the more conservative Instruct and Proprietary workers.
This complementarity further motivates our latent-variable formulation: diverse samples from heterogeneous workers collectively approximate a richer reasoning space than any single source, while the pair-aware scoring mechanism assigns higher weights to higher-quality samples.

We adopt a lenient threshold $\epsilon = -0.1$ across all worker types, allowing samples whose reasoning introduces only a small performance drop to be retained. 
As shown in Fig.~\ref{fig:gain_dist}, the Thinking worker produces the largest number of filtered samples, indicating that many of its reasoning chains substantially harm query--target alignment. 
Importantly, this does not imply that most samples are simply retained such that the selection mechanism becomes ineffective. 
Instead, the key aspect of our design lies in the relative weighting among the accepted samples, which is determined by the pair-aware alignment scores across different workers.

\subsection{Distribution of Reasoning Utility}\label{app:utility_dist}

\begin{figure}[t]
\centering
\includegraphics[width=\columnwidth]{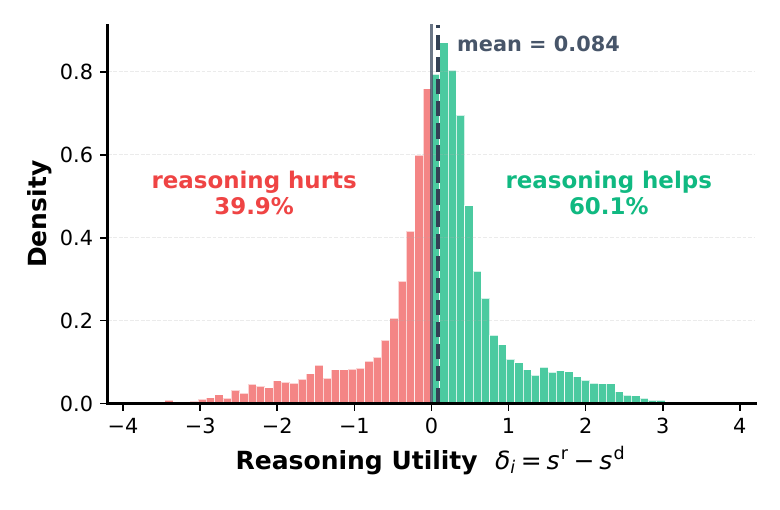}
\caption{Distribution of reasoning utility $\delta_i$ over the training set. Green bars ($\delta_i \geq 0$) indicate instances where reasoning improves retrieval; red bars ($\delta_i < 0$) indicate the opposite.}
\label{fig:utility_dist}
\end{figure}

As described in \textsection~\ref{sec:utility}, we compute the reasoning utility $\delta_i = s_i^{\mathrm{r}} - s_i^{\mathrm{d}}$ for each training query by comparing the normalized similarity scores obtained from reasoning-enhanced and direct embeddings.
Fig.~\ref{fig:utility_dist} shows the distribution of $\delta_i$ across a subset. The distribution is unimodal and centered slightly above zero, with a longer right tail than left.
Roughly 60\% of instances exhibit positive utility, confirming that reasoning-enhanced embeddings are generally beneficial after pair-aware selection.
However, a substantial 40\% of instances show negative utility, meaning that reasoning introduces noise or obscures salient signals for these samples.
This mixed landscape directly motivates our adaptive reasoning mechanism: a one-size-fits-all strategy---whether always reasoning or never reasoning---is inherently suboptimal.
The continuous, instance-dependent nature of $\delta_i$ further justifies using reinforcement learning rather than a hard threshold to learn the decision boundary, as the optimal reasoning policy must account for fine-grained variations across inputs.

\begin{table*}[ht!]

\begin{tcolorbox}[
    enhanced,
    segmentation style={solid, line width=0.8pt, color=black!40}
]

{\normalsize \textbf{==~\textsc{Instruction to Instruct-based and High-capacity Proprietary Models}~==}}

\medskip
The input above is a query or a candidate item for retrieval. Carefully analyze the input, which may contain text, images, videos, or any combination of modalities. 

Identify the core semantic content, including the main topic, key entities, relevant relationships, contextual information, and any salient details that are essential for understanding its meaning. 

Provide a concise, factual reasoning process that focuses on semantic understanding rather than speculation. Enclose this reasoning within \texttt{<reason>} and \texttt{</reason>} tags. Then, on a new line, produce a compact semantic summary that best represents the input for retrieval purposes. The summary should be either a single word or a short sentence, and must begin with \texttt{<sum>}.

Your response must strictly follow this format:\\
\texttt{<reason>}\\
\texttt{...}\\
\texttt{</reason>}\\
\texttt{<sum>...}\\
Any response that does not follow this format is invalid.

\tcbline

{\normalsize \textbf{==================~\textsc{Instruction to Thinking Models}~==================}}

\medskip
The above input is a query/candidate for retrieval. Carefully examine and analyze the above input (which may include text, images, videos, or any combination). Identify and describe the key elements present in the input, such as the main topic, important entities, relationships, context, and any notable features or details that contribute to the overall meaning. Finally, synthesize your analysis and reflection into a single word or a concise sentence that best captures the essence of the input for retrieval purposes. If the input is a phrase or word, the summary is that word itself.

\end{tcolorbox}
\caption{Instructions to different workers MLLM for prior distribution simulation.}
\label{tab:app_worker_inst}
\end{table*}

\subsection{Detailed Results on MMEB-V2 and MMEB-V1}\label{app:detailed_results}
 
We provide per-dataset results on MMEB-V2 (78 datasets, three modality groups) in Tab.~\ref{tab:app_detailed_score}.
For compatibility with prior work evaluated on the original image-only benchmark, we also report MMEB-V1 results (36 datasets) in Tab.~\ref{tab:app_v1_results}.
\modelname{} (Qwen3-VL-4B) achieves 74.8 overall on V1, outperforming all baselines including Embed-RL-4B and UME-R1-7B, confirming that the benefits of our framework are not specific to the video and document modalities newly introduced in V2.

To assess whether the performance improvements of our model are statistically significant across tasks, we conduct paired statistical tests based on the per-task scores reported in Table~\ref{tab:app_detailed_score}. We compare MMEmb-R1 (Qwen3-VL-4B) against several strong baselines using the Wilcoxon signed-rank test as the primary non-parametric test, a paired $t$-test as a complementary analysis, and task-level bootstrap 95\% confidence intervals with 10k resamples over the paired differences.As shown in Table~\ref{tab:statistical_significance}, MMEmb-R1 consistently outperforms the compared baselines across the 78 tasks. In particular, the improvements over the closest 4B rival, Embed-RL-4B, are observed on 69 out of 78 tasks, with a mean improvement of 3.08 points and a 95\% bootstrap confidence interval of $[2.13, 4.02]$. Both the Wilcoxon signed-rank test and the paired $t$-test yield highly significant $p$-values for all comparisons, while all corresponding confidence intervals remain strictly above zero. These results demonstrate that the improvements of MMEmb-R1 are statistically significant and robust across tasks.

\begin{table*}[t]
    \centering
    \small
    \setlength{\tabcolsep}{4pt}
    \begin{tabular}{lccccc}
        \toprule[1.5pt]
        \textbf{Comparison (Ours-4B vs.)} &
        \textbf{Win/Loss/Tie} &
        \textbf{Mean $\Delta$} &
        \textbf{95\% CI of $\Delta$} &
        \textbf{Wilcoxon $p$} &
        \textbf{Paired-$t$ $p$} \\
        \midrule
        Embed-RL-4B (closest 4B rival) &
        69 / 9 / 0 &
        +3.08 &
        [+2.13, +4.02] &
        $6.3 \times 10^{-10}$ &
        $1.5 \times 10^{-8}$ \\
        
        UME-R1-7B &
        63 / 13 / 2 &
        +6.66 &
        [+4.36, +9.09] &
        $4.4 \times 10^{-8}$ &
        $4.3 \times 10^{-7}$ \\
        
        Embed-RL-2B &
        69 / 9 / 0 &
        +4.39 &
        [+3.27, +5.52] &
        $3.4 \times 10^{-11}$ &
        $5.8 \times 10^{-11}$ \\
        \bottomrule[1.5pt]
    \end{tabular}
    \caption{Statistical significance analysis of MMEmb-R1 (Qwen3-VL-4B) against strong baselines over 78 tasks. $\Delta$ denotes the per-task performance difference between our model and the corresponding baseline.}
    \label{tab:statistical_significance}
\end{table*}

 \begin{table}[h]
\centering
\small
\begin{tabular}{lc}
\toprule[1.5pt]
Worker Configuration & MMEB-V2 \\
\midrule
InternVL3-2B (single) & 54.8 \\
GLM-4.1V-Thinking (single) & 61.2 \\
Qwen3VL-32B (single) & 61.9 \\
GLM-4.1V-Thinking + InternVL3-14B & 63.0 \\
\modelname{} (Full) & \textbf{65.0} \\
\modelname + Another 3 workers & 62.1 \\
\bottomrule[1.5pt]
\end{tabular}
\caption{MMEB-V2 performance under different worker configurations.}
\label{tab:app_worker_type}
\end{table}
 \begin{table}[h]
\centering
\small
\begin{tabular}{lc}
\toprule[1.5pt]
Model/Evaluator & MMEB-V2 \\
\midrule
UME-R1 & 60.1 \\
Evaluator-Flash & 63.2 \\
InternVL3-38B & 63.7 \\
Qwen3-VL-32B (Alt.\ Prompt) & 64.7 \\
Qwen3-VL-32B (Original) & 65.0 \\
\bottomrule[1.5pt]
\end{tabular}
\caption{MMEB-V2 performance under different evaluator configurations. ``Alt.\ Prompt'' denotes the use of an alternative prompt template.}
\label{tab:app_eval_type}
\end{table}

\subsection{Deep Analysis of Workers and Evaluator}

\add{
We employ a worker--evaluator framework to approximate the latent reasoning space. 
To assess the robustness of different worker and evaluator configurations, as well as to gain deeper insight into the proposed pair-aware reasoning selection mechanism, we conduct the following experiments. 
We consider different types of workers, as listed in Tab.~\ref{tab:app_worker_type}, and different evaluator configurations, as shown in Tab.~\ref{tab:app_eval_type}.
}

\add{
We observe the following findings:
\textbf{(1)} 
Basically, increasing the diversity and capability of worker models consistently improves downstream performance, supporting our hypothesis that heterogeneous reasoning candidates better approximate the latent reasoning space. However, continuously adding more heterogeneous workers may not always be beneficial. When we further add three additional workers (Kimi/Claude/GPT series), performance decreases, likely because excessive diversity introduces additional challenges for CoT-based training and makes the optimization process more difficult.
\textbf{(2)} 
We acknowledge that introducing multiple workers and a large evaluator increases engineering overhead. 
To reduce this cost, we replace the original 32B evaluator with a compact 2B reward model trained to mimic the original scoring process, denoted as \emph{Evaluator-Flash}. 
Despite its significantly smaller size, this variant achieves a score of 63.2, outperforming UME-R1 (60.1) while providing a more compact and scalable pipeline.
\textbf{(3)} 
The evaluation stage remains stable across different evaluator models and prompting strategies, provided that the evaluator is sufficiently capable. 
In contrast, worker quality is substantially more critical, since workers directly generate the reasoning data used for training. 
This observation further justifies our sophisticated pair-aware selection mechanism, which is specifically designed to filter and weight high-quality reasoning supervision.
\textbf{(4)} 
We further conduct an experiment using Qwen3-VL-32B as the worker model for generating CoT data. 
As shown in Tab.~\ref{tab:app_eval_type}, even with a substantially stronger teacher model, the resulting performance still falls short of our full framework. 
This suggests that the improvements mainly stem from the effectiveness of our proposed methodology, rather than simply distilling knowledge from more powerful models.
}

\subsection{Results Generalized to Other Domains}

\add{
To further demonstrate the generalizability of our framework, we conduct preliminary experiments on two additional benchmarks. 
First, we evaluate on MMStar~\cite{chen2024we}, a widely used multimodal benchmark, to examine whether our pair-aware reasoning mechanism can transfer to generative multimodal tasks. 
Second, we evaluate on ROCOv2~\cite{ruckert2024rocov2}, an out-of-domain medical multimodal dataset. 
We compare model performance with and without the proposed pair-aware selection mechanism. 
As shown in Tab.~\ref{tab:app_gene}, our method consistently improves performance across both settings, suggesting that the proposed framework generalizes effectively beyond the primary benchmark.
}

\begin{table}[h]
\centering
\small
\begin{tabular}{lcc}
\toprule[1.5pt]
Benchmark & w/o Selection & w/ Pair-Aware Selection \\
\midrule
MMStar & 60.2 & \textbf{63.2} \\
ROCOv2 & 43.26 & \textbf{45.78} \\
\bottomrule[1.5pt]
\end{tabular}
\caption{Performance with and without the pair-aware selection mechanism on a generative task (MMStar) and an out-of-domain benchmark (ROCOv2). ROCOv2 is evaluated using CUI@5.}
\label{tab:app_gene}
\end{table}

\subsection{Query-Dependent Embeddings}

Our embedding is designed to capture not only the semantics of the visual and textual content, but also the question or instruction associated with the query. To verify this property, we conduct controlled experiments on MMEB-V2 by perturbing the question text while keeping the target items unchanged. Specifically, we consider two settings: (1) \textit{w/o question text}, where the question text is removed from every query, and (2) \textit{random question text}, where the question text is randomly reassigned across queries. All other components of the queries and the target items remain unchanged.
The results are shown in Tab.~\ref{tab:app_queries}. Removing the question text causes a substantial performance drop from 65.0 to 52.8, indicating that the question text provides essential information for constructing discriminative query embeddings. Moreover, randomly replacing the question text further degrades the performance to 43.1, suggesting that the embedding is sensitive to the specific query intent rather than merely exploiting generic textual information. These results shows that our model produces query-dependent embeddings that adapt to the question or instruction associated with each query.

\begin{table}[t]
    \centering
    \small
    \begin{tabular}{lc}
        \toprule[1.5pt]
        \textbf{Setting} & \textbf{MMEB-V2} \\
        \midrule
        Original & 65.0 \\
        w/o question text & 52.8 \\
        Random question text & 43.1 \\
        \bottomrule[1.5pt]
    \end{tabular}
    \caption{MMEB-V2 performance under different question-text perturbations. The target items remain unchanged across all settings.}
    \label{tab:app_queries}
\end{table}
\section{More Implementation Details}\label{sec:app_impl}
\subsection{Multi-Worker Reasoning Path Generation}\label{sec:app_worker_prompt}
As discussed in \textsection~\ref{sec:candidate_gen}, we leverage heterogeneous MLLMs to approximate the distribution of the reasoning latent space.
The key principle is to maximize complementarity across workers: each type contributes distinct reasoning styles and knowledge coverage, collectively simulating a richer prior than any single model.
Specifically, we employ the following models with carefully designed prompts, inspired by UME-R1~\cite{ume} and TTE~\cite{tte}.

\paragraph{Instruct-based models.}
We use InternVL3-14B-Instruct~\cite{zhu2025internvl3exploringadvancedtraining} with the prompt shown in Tab.~\ref{tab:app_worker_inst} (top).
The prompt enforces a structured \texttt{<reason>}...\texttt{<sum>} format, encouraging concise, factual semantic analysis.
Empirically, this worker produces the most consistently formatted and retrieval-oriented rationales, serving as a stable baseline in the candidate pool.

\paragraph{Thinking models.}
We leverage GLM-4.1V-Thinking~\cite{vteam2026glm45vglm41vthinkingversatilemultimodal}, using the prompt adopted in UME-R1~\cite{ume} (Tab.~\ref{tab:app_worker_inst}, bottom).
Unlike the instruct-based prompt, we do not enforce a rigid output format, allowing the model to reason in its native chain-of-thought style.
This produces longer, more exploratory chains with higher variance---occasionally yielding uniquely high-gain candidates, as shown in Appendix~\ref{app:gain_dist}.

\paragraph{High-capacity proprietary models.}
We further leverage the API of Doubao-Seed-1.6-Vision~\cite{seed1.6}, using the same prompt template as the instruct-based models.
Despite sharing the prompt, the proprietary model generates qualitatively richer rationales due to its broader world knowledge, resulting in the highest median reasoning gain among all worker types.

\begin{table*}[ht!]

\begin{tcolorbox}[
    enhanced,
    segmentation style={solid, line width=0.8pt, color=black!40}
]
{\normalsize \textbf{=================~\textsc{Baseline Prompt (without rationale)}~=================}}

\medskip

Above are two items for a retrieval task:

\medskip

\textbf{Query Item:}\\
\texttt{<image><video>} \\
Text: \texttt{\{query\_text\}}

\medskip

\textbf{Target Item:}\\
\texttt{<image><video>} \\
Text: \texttt{\{target\_text\}}

\medskip

Based on the query item and target item, are these two items semantically relevant for retrieval? Consider whether they match in terms of content, topic, and intent.

\medskip

Answer with \texttt{YES} or \texttt{NO} only.

\tcblower

{\normalsize \textbf{========~\textsc{With-Rationale Prompt (for counterfactual evaluation)}~========}}

\medskip

Above are two items for a retrieval task:

\medskip

\textbf{Query Item:}\\
\texttt{<image><video>} \\
Text: \texttt{\{query\_text\}}\\
Rationale: \texttt{\{query\_rationale\}}

\medskip

\textbf{Target Item:}\\
\texttt{<image><video>} \\
Text: \texttt{\{target\_text\}}\\
Rationale: \texttt{\{target\_rationale\}}

\medskip

The rationales are reasoning paths generated to help understand the semantic content of each item. Based on the query item, target item, and their rationales, does adding the rationale improve the retrieval effectiveness? Consider whether the rationales capture essential semantic information that aids in matching these two items.

\medskip

Answer with \texttt{YES} or \texttt{NO} only.

\end{tcolorbox}
\caption{Prompts used for the pair-aware evaluator $\mathcal{J}$. The \emph{baseline} prompt (top) evaluates query--target relevance without reasoning. The \emph{with-rationale} prompt (bottom) includes generated rationales, enabling the counterfactual comparison $\Delta_r = c_r - c_0$.}
\label{tab:app_evaluator_inst}
\end{table*}

\subsection{Pair-Aware Evaluator Implementation}\label{sec:app_evaluator}

For the pair-aware evaluator $\mathcal{J}$, we employ Qwen3-VL-32B-Instruct~\cite{yang2025qwen3technicalreport} and use vLLM\footnote{\url{https://vllm.ai/}} for efficient inference. We leverage this strong open-source model to obtain reliable logit scores for relevance estimation. 
The evaluator prompts are provided in Tab.~\ref{tab:app_evaluator_inst}.
For each training pair $(q_i, t_i^+)$ and each reasoning candidate $r \in \mathcal{R}_i$ generated by the worker models, we perform two inference passes through the evaluator.
In the \emph{baseline pass}, the evaluator receives only the raw query and target (with their associated images or videos) and is prompted to judge semantic relevance with a binary YES/NO response.
In the \emph{with-rationale pass}, the same query--target pair is augmented with the candidate rationale.
We extract the logit of the first generated token for both \texttt{[YES]} and \texttt{[NO]}, and compute a log-probability ratio $\text{diff} = \log p(\texttt{[YES]}) - \log p(\texttt{[NO]})$ for each pass.
The ``counterfactual reasoning gain'' is then:
$
\Delta_r = \text{diff}_{\text{with}} - \text{diff}_{\text{baseline}}.
$
A positive $\Delta_r$ indicates that the rationale improves the evaluator's confidence in the query--target match beyond what the raw inputs alone provide.
After computing $\Delta_r$ for all three worker sources (Instruct, Thinking, Proprietary), we apply a softmax over the scores to obtain normalized selection weights:
$
w_k = \text{softmax}(\Delta_{r_k})$, $k \in \{\text{instruct}, \text{thinking}, \text{proprietary}\}.
$
These weights are stored alongside the rationales and used during training for weighted sampling.

Moreover, we notice that the prompts used in the baseline and rationale conditions are slightly different, raising a concern that the observed differences may be partially attributed to prompt-induced bias. To address this concern, we redesigned the evaluator prompts such that both conditions ask exactly the same relevance question (“Given the query and target, are they relevant for retrieval? YES/NO”), with the only difference being whether the rationale is included in the evaluator context. Using this matched-prompt setting, we recomputed all per-worker selection weights and compared them with the original ones. The resulting rankings show high agreement, with Kendall's $\tau=0.82$, indicating that the original weights are not primarily driven by prompt asymmetry. Moreover, the updated 2B model trained with the revised weights achieves 64.8 on MMEB-V2, comparable to the original setting. Therefore, we retain the original prompt for simplicity.

\begin{table*}[ht!]

\begin{tcolorbox}[
    enhanced,
    segmentation style={solid, line width=0.8pt, color=black!40}
]
{
\texttt{<input\_image>}\texttt{<input\_video>}\texttt{<input\_text>}\texttt{<d\_emb>} Represent the above input text, images, videos, or any combination of the three as embeddings. You may output the thinking process in \texttt{<reason>} \texttt{</reason>} tags and then summarize the entire input in a word or sentence. Finally, use the \texttt{<r\_emb>} tag to represent the entire input. If explicit reasoning is not necessary (e.g., the task is simple or the input is concise), you may directly produce the embeddings without generating intermediate thinking.
}
\end{tcolorbox}
\caption{Instruction template used during joint reasoning and embedding training (\textsection~\ref{sec:joint_training}).}
\label{tab:app_sft_inst}
\end{table*}

\subsection{Details of Joint Reasoning and Embedding Training}\label{app:joint_training}

We adopt the instruction template shown in Tab.~\ref{tab:app_sft_inst}.
Two special tokens are introduced: \texttt{<d\_emb>}, appended at the beginning of the instruction to mark the direct embedding extraction point, and \texttt{<r\_emb>}, generated after optional reasoning tokens to mark the reasoning-enhanced embedding extraction point.
Although the adaptive reasoning policy is formally learned in the RL stage (\textsection~\ref{sec:adaptive_rl}), we find it beneficial to expose the model to a small fraction of direct-embedding samples during joint training, preventing the model from becoming overly reliant on reasoning generation and easing the subsequent policy learning.
Specifically, we select samples that are unlikely to benefit from reasoning based on two criteria: (1) samples with very low pair-aware selection weight $w_r$, indicating that no generated rationale meaningfully improves query--target alignment, and (2) samples with very short input text (fewer than 5 words), where reasoning would constitute overthinking.
For these samples, we replace the rationale with an \texttt{<empty>} token with probability 0.1, training the model to directly produce embeddings without intermediate reasoning.

During training, the vision encoder is kept frozen, while both the multimodal projector and the LLM backbone are updated.
We train for 3 epochs with a per-device batch size of 4 and gradient accumulation steps of 8, yielding an effective batch size of 256 across 8 GPUs.
We use AdamW with a learning rate of $5 \times 10^{-5}$, cosine scheduling, and a warmup ratio of 0.03.
The loss weights $\lambda_{\text{CoT}}$ and $\lambda_{\text{direct}}$ are both set to 1.
The maximum sequence length is 12288 tokens, with image pixels clipped to $[768,\, 2359296]$.
Training is conducted in \texttt{bfloat16} precision with DeepSpeed ZeRO-3 and gradient checkpointing.

\subsection{Details of Adaptive Reasoning Control}\label{app:adaptive}

For adaptive reasoning control, we adopt the codebase of VLM-R1\footnote{https://github.com/om-ai-lab/VLM-R1}~\cite{shen2025vlmr1stablegeneralizabler1style}. 
We sample $8$ completions for each query with a maximum generation length of $1024$ and temperature $1.0$. 
The GRPO clipping coefficient is set to the range $[0.8, 1.28]$, and the KL-divergence coefficient is set to $0.04$. 
Training is performed with a batch size of $8$ per device and $2$ gradient accumulation steps. 
The learning rate is $1\times10^{-6}$, and the model is trained for $2$ epochs. Additional details regarding GRPO are provided in Section~\ref{sec:grpo}.

For the adaptive reward design, we set $\alpha = 0.2$ and encourage the \textsc{Direct} action during the first $500$ training steps. 
The reasoning cost coefficient is set to $c = 1\times10^{-3}$. 
For the format reward, any chain-of-thought (CoT) that deviates from our predefined format receives a reward of $0$, while valid outputs receive a reward of $1$.
For the embedding reward, we adopt the design proposed in UME-R1~\cite{ume}, which measures how well the generated representations distinguish positive targets from negative ones. 
Specifically, the reward considers two criteria: 
(i) the ranking of positive targets among negative targets, and 
(ii) the similarity gap between positives and negatives.
For each query $q$ with a positive target $t^{+}$ and a negative target $t^{-}$, we sample a group of responses 
$\{o^{+}_{j}\}_{j=1}^{G}$ associated with the positive target and another group 
$\{o^{-}_{j}\}_{j=1}^{G}$ associated with the negative target. 
Given the $i$-th sampled response $o_i$ and embedding model $\mathcal{E}_\theta$, we compute its similarity scores with the positive targets as $
S^{+} = \{\mathcal{E}_\theta([q, o_i])\cdot \mathcal{E}_\theta([t^{+}, o^{+}_{j})]\}_{j=1}^{G}
$
and with the negative targets as
$
S^{-} = \{\pi_\theta(q, o_i)\cdot \pi_\theta(t^{-}, o^{-}_{j})\}_{j=1}^{G}.
$
The embedding reward is defined as:
\begin{equation*}
\begin{aligned}
R_{\text{emb}}(o_i) =&
\frac{\left|S^{+} \cap \text{top}_{G}(S^{+}\cup S^{-})\right|}{G}\\
& \times\left(\text{avg}(S^{+}) - \text{avg}(S^{-})\right),
\end{aligned}
\end{equation*}
where $\text{top}_{G}(\cdot)$ selects the $G$ largest elements from the input set. 
The first term measures whether positive similarities rank higher than negative ones, while the second term captures the magnitude of the similarity gap. 
Maximizing this reward encourages the model to produce reasoning trajectories that lead to more discriminative and informative embeddings.
Moreover, we treat the query and the target symmetrically and compute the final reward as the mean of the rewards obtained from both directions.

\subsection{Discussion on the Implementation of Other Baselines}

\add{We provide additional implementation details of existing reasoning-driven embedding methods to facilitate a clearer comparison with our framework. 
TTE employs Qwen2.5-VL-72B as the teacher model and designs dataset-specific prompts to generate CoT annotations. 
UME-R1 uses GLM-4.1V-Thinking for rationale generation followed by heuristic filtering, while EMB-RL relies on teacher-generated CoT data together with model-based filtering strategies. 
In terms of training pipelines, both UME-R1 and EMB-RL adopt a two-stage supervised fine-tuning (SFT) plus reinforcement learning (RL) framework, whereas TTE separately trains the embedder and the reasoner. 
Therefore, the overall complexity of our framework is comparable to existing baselines rather than substantially higher. 
Moreover, as the data scale increases, our method may even become more efficient by avoiding dataset-specific manual prompt engineering and heuristic filtering procedures.}\par

\section{Background and Preliminaries}\label{app:background}

\subsection{Causal Inference and Causal Learning}\label{app:causal}

Causal inference~\cite{pearl2009causality} aims to identify cause-and-effect relationships beyond associational patterns.
The structural causal model (SCM) framework represents data-generating processes as directed acyclic graphs, where Pearl's do-operator $\mathrm{do}(X=x)$ formalizes interventions by fixing a variable while severing its incoming causal edges.
This distinguishes the interventional distribution $P(Y\mid\mathrm{do}(X\!=\!x))$ from the observational conditional $P(Y\mid X\!=\!x)$, enabling isolation of the true causal effect from confounders.
A key quantity is the average treatment effect: $\text{ATE} = \mathbb{E}[Y \mid \mathrm{do}(X\!=\!1)] - \mathbb{E}[Y \mid \mathrm{do}(X\!=\!0)]$.
At a higher level, counterfactual reasoning~\cite{pearl2018book} addresses ``what if'' questions---computing the outcome under an alternative intervention for the same instance.
Causal perspectives have been increasingly adopted in the deep learning community~\cite{yang2023context,yang2024towards}.
These works share a common principle: explicitly modeling causal pathways isolates target effects from confounders, yielding more robust systems. In our work, obtaining an ideal reasoning trace is challenging. From a Bayesian perspective, the diverse rationales generated by multiple MLLMs can be viewed as a noisy prior candidate set, and the evaluator-based reweighting process provides a more informative posterior distribution. During evaluation, the evaluator estimates whether introducing a reasoning path improves retrieval relevance and quantifies its contribution. The comparison between predictions with and without rationale is conceptually related to the intuition of comparing factual and counterfactual outcomes in causal analysis. 

\subsection{Group Relative Policy Optimization}\label{app:grpo}

Standard RLHF alignment via PPO~\cite{schulman2017proximal} requires a separate critic network to estimate per-token advantages, introducing substantial memory overhead when the policy is a large language model. 
An alternative line of work replaces reinforcement learning with preference-based optimization. 
Direct Preference Optimization (DPO)~\cite{rafailov2023direct} learns from paired preference data $(x, y^+, y^-)$ by directly optimizing the likelihood difference between preferred and rejected responses, eliminating the need for an explicit reward model or policy gradient updates. 
However, DPO relies on curated preference pairs~\cite{xu2024dpo} and does not naturally accommodate reward signals derived from multiple sampled outputs.
GRPO~\cite{shao2024deepseekmathpushinglimitsmathematical} addresses these limitations by computing advantages at the group level.  For each input $x$, $G$ candidate outputs $\{o_1, \dots, o_G\}$ are sampled from $\pi_\theta$ and scored by a reward function $R(\cdot)$, with advantages normalized within the group:
$
\hat{A}_i = ({R(o_i) - \mathrm{mean}(\{R(o_j)\})}) / {\mathrm{std}(\{R(o_j)\})}.
$
The policy is updated via a clipped surrogate objective with KL regularization against a reference policy $\pi_{\mathrm{ref}}$:
\begin{equation*}
\begin{aligned}
\mathcal{L}_{\text{GRPO}} 
&= \mathbb{E}\Bigl[\min\bigl(r_i\hat{A}_i,\; 
\mathrm{clip}(r_i, 1-\epsilon, 1+\epsilon)\hat{A}_i\bigr)\Bigr] \\
&\quad - \beta\, D_{\mathrm{KL}}(\pi_\theta \| \pi_{\mathrm{ref}})
\end{aligned}
\end{equation*}
where $r_i = \pi_\theta(o_i \mid x) / \pi_{\mathrm{old}}(o_i \mid x)$.
GRPO has been widely adopted in reasoning model training, notably by DeepSeek-R1~\cite{Guo_2025}.
Its key advantages are: no critic network (lower memory), stable gradients via group normalization, and straightforward implementation atop standard LM training infrastructure~\cite{liu2025reinforcement}.

\begin{table*}[htbp]
    \tabcolsep=0.2cm
    \centering
    \caption{Detailed results of baselines and \modelname{} on full MMEB v2 benchmark. Rows are colored by modality: \colorbox{blue!4}{Image}, \colorbox{orange!5}{Video}, and \colorbox{green!5}{VisDoc}. Best results per row are in \textbf{bold}.}
    \renewcommand{\arraystretch}{1.1}
    \begin{adjustbox}{width=\textwidth}
    \begin{tabular}{l|cccccccccccc}
        \toprule
        ~ & ColPali v1.3 & GME-7B & VLM2Vec-7B & VLM2Vec-V2-2B & CAFe-7B & UME-R1-2B & UME-R1-7B & Embed-RL-2B & Embed-RL-4B & Ours (Qwen2-VL-2B) & Ours (Qwen3-VL-4B) \\ 
        \midrule

        \rowcolor{blue!4}
        ImageNet-1K & 42.4 & 64.6 & 80.1 & 80.8 & 77.3 & 75.3 & 80.4 & 78.0 & 79.5 & 76.2 & \textbf{81.8} \\ 
        \rowcolor{blue!4}
        N24News & 25.5 & 50.5 & 79.7 & 72.9 & \textbf{83.2} & 81.1 & 82.3 & 44.9 & 48.3 & 56.5 & 62.2 \\ 
        \rowcolor{blue!4}
        HatefulMemes & 50.6 & 53.6 & 69.7 & 56.3 & 78.7 & 75.2 & \textbf{79.0} & 65.0 & 66.2 & 71.8 & 70.5 \\ 
        \rowcolor{blue!4}
        VOC2007 & 69.8 & 80.3 & 80.7 & 85.0 & 89.8 & 80.0 & \textbf{90.8} & 78.7 & 79.5 & 79.5 & 82.1 \\ 
        \rowcolor{blue!4}
        SUN397 & 56.1 & 69.5 & 77.4 & 71.0 & 79.9 & 79.4 & \textbf{80.3} & 75.4 & 79.2 & 76.2 & 79.7 \\ 
        \rowcolor{blue!4}
        Place365 & 27.5 & 39.1 & 37.4 & 35.9 & 45.0 & 42.6 & 46.8 & 43.9 & 43.1 & \textbf{48.5} & 46.8 \\ 
        \rowcolor{blue!4}
        ImageNet-A & 14.9 & 41.2 & 58.1 & 47.4 & 55.2 & 50.4 & 53.9 & 59.2 & 58.1 & 53.8 & \textbf{62.3} \\ 
        \rowcolor{blue!4}
        ImageNet-R & 64.6 & 83.9 & 73.9 & 89.3 & 88.0 & 88.7 & 90.1 & 88.5 & 88.2 & 89.2 & \textbf{91.5} \\ 
        \rowcolor{blue!4}
        ObjectNet & 45.6 & 69.0 & 40.1 & 65.2 & 22.5 & 52.0 & 42.3 & 74.8 & 75.4 & 62.5 & \textbf{75.9} \\ 
        \rowcolor{blue!4}
        Country211 & 6.0 & 24.8 & 29.8 & \textbf{30.2} & 16.7 & 23.4 & 25.0 & 20.0 & 19.4 & 25.5 & 23.8 \\ 
        \rowcolor{blue!4}
        OK-VQA & 9.4 & 33.2 & 56.8 & 51.5 & 67.3 & 62.4 & 71.7 & 61.4 & 67.3 & 66.8 & \textbf{68.9} \\ 
        \rowcolor{blue!4}
        A-OKVQA & 6.6 & 21.0 & 47.3 & 43.6 & 63.8 & 51.1 & 58.7 & 54.7 & 59.3 & 57.2 & \textbf{64.1} \\ 
        \rowcolor{blue!4}
        DocVQA & 11.3 & 41.4 & 89.7 & 90.1 & 79.2 & 92.2 & 93.8 & 92.4 & 94.3 & 92.8 & \textbf{95.2} \\ 
        \rowcolor{blue!4}
        InfographicsVQA & 5.0 & 20.3 & 60.0 & 58.8 & 53.3 & 67.7 & 79.2 & 76.7 & 77.5 & 78.5 & \textbf{82.1} \\ 
        \rowcolor{blue!4}
        ChartQA & 5.7 & 17.8 & 56.9 & 47.4 & 48.8 & 64.9 & 75.1 & 80.7 & 80.9 & 75.2 & \textbf{81.5} \\ 
        \rowcolor{blue!4}
        Visual7W & 6.1 & 22.2 & 52.7 & 52.9 & 52.5 & 54.1 & 55.2 & 52.7 & 55.3 & 58.0 & \textbf{65.2} \\ 
        \rowcolor{blue!4}
        ScienceQA & 16.3 & 28.0 & 38.5 & 38.2 & \textbf{65.4} & 42.7 & 53.7 & 57.3 & 61.6 & 49.8 & 64.8 \\ 
        \rowcolor{blue!4}
        VizWiz & 27.6 & 39.0 & 39.9 & 43.3 & 43.8 & 46.8 & 51.6 & 54.5 & 56.2 & 54.5 & \textbf{64.5} \\ 
        \rowcolor{blue!4}
        GQA & 8.3 & \textbf{76.9} & 55.1 & 64.9 & 65.7 & 67.3 & 69.3 & 64.9 & 68.5 & 64.2 & 69.8 \\ 
        \rowcolor{blue!4}
        TextVQA & 18.8 & 46.8 & 71.6 & 72.2 & 76.8 & 78.6 & 83.5 & 83.8 & 84.3 & 83.5 & \textbf{85.9} \\ 
        \rowcolor{blue!4}
        VisDial & 41.2 & 60.8 & 81.9 & 82.7 & 82.7 & 76.6 & 80.7 & 81.5 & \textbf{84.9} & 83.2 & 83.5 \\ 
        \rowcolor{blue!4}
        CIRR & 8.2 & 54.9 & 51.1 & 57.5 & 60.4 & 53.7 & 55.3 & 47.6 & 61.2 & 52.8 & \textbf{69.7} \\ 
        \rowcolor{blue!4}
        VisualNews\_t2i & 50.1 & \textbf{79.7} & 80.5 & 74.5 & 69.5 & 71.7 & 76.8 & 71.9 & 73.7 & 63.5 & 75.1 \\ 
        \rowcolor{blue!4}
        VisualNews\_i2t & 47.6 & \textbf{83.6} & 81.2 & 78.2 & 79.4 & 74.2 & 82.0 & 73.6 & 73.9 & 74.8 & 77.3 \\ 
        \rowcolor{blue!4}
        MSCOCO\_t2i & 59.2 & 71.2 & 77.2 & 75.3 & 75.4 & 75.1 & 78.3 & 79.4 & 78.9 & 81.2 & \textbf{84.5} \\ 
        \rowcolor{blue!4}
        MSCOCO\_i2t & 49.9 & 57.7 & 73.9 & 71.4 & 73.1 & 68.9 & 71.4 & 75.3 & \textbf{76.3} & 73.5 & 76.2 \\ 
        \rowcolor{blue!4}
        NIGHTS & 65.5 & 67.6 & 67.6 & 68.6 & 66.7 & 67.2 & 68.1 & 66.3 & 66.4 & \textbf{73.2} & 70.9 \\ 
        \rowcolor{blue!4}
        WebQA & 53.8 & \textbf{91.4} & 88.3 & 90.6 & 89.3 & 90.0 & 90.9 & 89.3 & 90.5 & 87.8 & 91.2 \\ 
        \rowcolor{blue!4}
        FashionIQ & 5.9 & \textbf{37.8} & 17.1 & 19.5 & 39.0 & 17.1 & 23.4 & 24.0 & 31.9 & 26.5 & \textbf{37.8} \\ 
        \rowcolor{blue!4}
        Wiki-SS-NQ & \textbf{80.5} & 78.2 & 62.3 & 66.9 & 61.2 & 62.0 & 72.5 & 68.9 & 69.6 & 70.2 & 74.1 \\ 
        \rowcolor{blue!4}
        OVEN & 50.0 & \textbf{75.1} & 66.5 & 64.3 & 60.8 & 66.9 & 71.4 & 61.4 & 60.7 & 66.8 & 61.5 \\ 
        \rowcolor{blue!4}
        EDIS & 64.7 & \textbf{96.0} & 85.7 & 84.1 & 71.3 & 88.0 & 92.0 & 84.5 & 87.4 & 86.5 & 91.8 \\ 
        \rowcolor{blue!4}
        MSCOCO & 36.7 & 31.4 & 75.7 & 67.1 & 84.7 & 69.5 & 72.7 & 92.9 & 93.6 & 87.2 & \textbf{94.2} \\ 
        \rowcolor{blue!4}
        RefCOCO & 64.5 & 60.9 & 87.6 & 87.1 & 89.4 & 83.3 & 91.4 & 94.9 & 95.9 & 97.5 & \textbf{99.1} \\ 
        \rowcolor{blue!4}
        RefCOCO-Matching & 3.9 & 78.4 & 84.6 & 85.8 & 83.0 & 84.4 & 91.1 & 85.8 & 88.0 & 83.8 & \textbf{92.5} \\ 
        \rowcolor{blue!4}
        Visual7W-Pointing & 56.1 & 66.5 & 81.0 & 69.2 &93.2 & 71.5 & 84.2 & 88.0 & 87.9 & 87.2 & \textbf{93.8} \\ 

        \midrule

        \rowcolor{orange!5}
        K700 & 23.4 & 39.7 & 35.5 & 38.0 & 40.1 & 35.8 & 42.8 & 55.8 & 56.8 & 57.2 & \textbf{57.3} \\ 
        \rowcolor{orange!5}
        SmthSmthV2 & 25.1 & 30.6 & 32.1 & 42.8 & 35.8 & 44.1 & 50.4 & 56.7 & 59.5 & 55.8 & \textbf{64.8} \\ 
        \rowcolor{orange!5}
        HMDB51 & 24.8 & 47.9 & 42.2 & 40.9 & 46.9 & 54.4 & 58.3 & 56.7 & 60.1 & 61.2 & \textbf{62.5} \\ 
        \rowcolor{orange!5}
        UCF101 & 49.4 & 54.7 & 61.8 & 60.0 & 39.6 & 67.2 & 70.0 & 79.3 & 78.5 & 74.5 & \textbf{81.2} \\ 
        \rowcolor{orange!5}
        Breakfast & 10.9 & 14.3 & 23.8 & 14.8 & 16.6 & 20.1 & 21.5 & 36.7 & 33.0 & 32.8 & \textbf{37.5} \\ 
        \rowcolor{orange!5}
        MVBench & 33.7 & 46.6 & 28.5 & 33.7 & 48.9 & 49.9 & \textbf{58.2} & 50.8 & 55.9 & 55.5 & \textbf{58.2} \\ 
        \rowcolor{orange!5}
        Video-MME & 30.6 & 39.2 & 27.8 & 30.7 & 46.0 & 41.7 & 47.3 & 47.1 & 50.5 & 49.8 & \textbf{52.1} \\ 
        \rowcolor{orange!5}
        NExTQA & 35.2 & 53.6 & 20.3 & 20.9 & 62.4 & 59.9 & \textbf{69.6} & 53.9 & 58.2 & 53.2 & 60.8 \\ 
        \rowcolor{orange!5}
        EgoSchema & 38.4 & 46.8 & 21.8 & 34.0 & \textbf{60.0} & 45.4 & 52.4 & 53.0 & 52.8 & 37.5 & 57.2 \\ 
        \rowcolor{orange!5}
        ActivityNetQA & 51.3 & 65.6 & 51.4 & 52.3 & 76.0 & 57.8 & 76.0 & 74.8 & 74.4 & 67.8 & \textbf{77.1} \\ 
        \rowcolor{orange!5}
        DiDeMo & 22.8 & 26.4 & 29.3 & 30.4 & 37.8 & 32.4 & 40.0 & 45.3 & 46.8 & 48.5 & \textbf{55.3} \\ 
        \rowcolor{orange!5}
        MSR-VTT & 17.6 & 31.8 & 34.5 & 28.3 & 36.5 & 34.3 & 38.9 & 45.7 & 46.2 & 47.2 & \textbf{51.8} \\ 
        \rowcolor{orange!5}
        MSVD & 45.4 & 49.7 & 46.7 & 48.1 & 56.4 & 55.4 & 60.8 & 67.2 & 65.8 & 62.8 & \textbf{68.2} \\ 
        \rowcolor{orange!5}
        VATEX & 16.7 & 24.9 & 25.5 & 26.5 & 32.0 & 29.9 & 32.6 & 43.6 & 43.4 & 44.2 & \textbf{52.9} \\ 
        \rowcolor{orange!5}
        YouCook2 & 5.3 & 9.1 & 9.0 & 10.6 & 9.5 & 12.7 & 18.5 & 23.5 & 23.3 & \textbf{35.5} & 29.7 \\ 
        \rowcolor{orange!5}
        QVHighlight & 19.9 & 59.5 & 57.7 & 49.4 & 58.4 & 57.5 & 54.9 & 70.7 & \textbf{73.6} & 48.2 & 68.3 \\ 
        \rowcolor{orange!5}
        Charades-STA & 29.0 & 14.0 & 19.8 & 20.2 & 18.7 & 20.4 & 21.9 & 26.4 & 25.0 & \textbf{34.8} & 31.5 \\ 
        \rowcolor{orange!5}
        MomentSeeker & 27.6 & 37.4 & 39.3 & 40.8 & 41.4 & 41.2 & 41.1 & 50.9 & 49.9 & 44.5 & \textbf{52.1} \\ 
        
        \midrule

        \rowcolor{green!5}
        ViDoRe\_arxivqa & 81.7 & 86.9 & 60.2 & 80.6 & 73.3 & 73.9 & 73.6 & 86.1 & 88.7 & 68.2 & \textbf{89.3} \\ 
        \rowcolor{green!5}
        ViDoRe\_docvqa & 56.6 & 57.5 & 34.7 & 44.9 & 38.3 & 37.9 & 41.1 & 45.7 & 47.5 & 46.8 & \textbf{53.8} \\ 
        \rowcolor{green!5}
        ViDoRe\_infovqa & 84.9 & \textbf{91.6} & 70.4 & 83.7 & 80.6 & 76.2 & 80.8 & 86.8 & 86.9 & 82.5 & 87.2 \\ 
        \rowcolor{green!5}
        ViDoRe\_tabfquad & 86.9 & \textbf{94.6} & 78.2 & 89.2 & 80.7 & 86.1 & 90.2 & 94.5 & 94.7 & 91.2 & 94.1 \\ 
        \rowcolor{green!5}
        ViDoRe\_tatdqa & 70.9 & \textbf{74.1} & 27.6 & 43.8 & 37.8 & 40.6 & 46.7 & 54.6 & 54.8 & 43.5 & 72.3 \\ 
        \rowcolor{green!5}
        ViDoRe\_shiftproject & 75.1 & \textbf{96.8} & 38.6 & 60.8 & 52.0 & 66.8 & 65.0 & 70.7 & 69.0 & 67.8 & 69.5 \\ 
        \rowcolor{green!5}
        ViDoRe\_artificial\_intelligence & 95.7 & \textbf{99.6} & 67.7 & 88.5 & 86.0 & 85.9 & 89.5 & 94.0 & 91.6 & 89.2 & 92.1 \\ 
        \rowcolor{green!5}
        ViDoRe\_energy & 94.7 & \textbf{95.3} & 60.4 & 86.5 & 84.8 & 83.3 & 85.7 & 86.7 & 88.1 & 81.5 & 88.6 \\ 
        \rowcolor{green!5}
        ViDoRe\_government\_reports & 93.6 & \textbf{98.8} & 61.8 & 85.0 & 85.0 & 82.6 & 89.8 & 89.0 & 90.7 & 84.8 & 91.2 \\ 
        \rowcolor{green!5}
        ViDoRe\_healthcare\_industry & 95.9 & \textbf{99.3} & 69.9 & 92.2 & 88.4 & 90.8 & 94.3 & 91.1 & 90.4 & 85.8 & 91.8 \\ 
        \rowcolor{green!5}
        ViDoRe\_esg\_reports\_human\_labeled\_v2 & 51.3 & 63.4 & 6.8 & 45.6 & 50.7 & 50.2 & 50.4 & 56.9 & 59.8 & 56.2 & \textbf{67.5} \\ 
        \rowcolor{green!5}
        ViDoRe\_biomedical\_lectures\_v2\_multilingual & 54.7 & 49.5 & 5.1 & 44.3 & 50.9 & 46.2 & 50.7 & 51.0 & 50.1 & 47.5 & \textbf{55.7} \\ 
        \rowcolor{green!5}
        ViDoRe\_economics\_reports\_v2\_multilingual & 49.0 & 54.2 & 13.9 & 43.0 & 54.3 & 45.7 & 57.8 & 53.0 & 53.9 & 59.2 & \textbf{64.3} \\ 
        \rowcolor{green!5}
        ViDoRe\_esg\_reports\_v2\_multilingual & 52.9 & 55.4 & 11.9 & 46.6 & 42.3 & 42.7 & 43.2 & 46.9 & 49.7 & \textbf{61.2} & 54.9 \\ 
        \rowcolor{green!5}
        VisRAG\_ArxivQA & 80.9 & 87.4 & 52.6 & 76.9 & 74.0 & 74.3 & 80.5 & 84.9 & 86.9 & 84.8 & \textbf{87.2} \\ 
        \rowcolor{green!5}
        VisRAG\_ChartQA & 72.3 & 86.1 & 57.7 & 83.7 & 82.7 & 86.0 & 85.0 & 88.3 & \textbf{88.5} & 74.2 & 88.1 \\ 
        \rowcolor{green!5}
        VisRAG\_MP-DocVQA & 82.0 & \textbf{89.7} & 60.6 & 88.1 & 75.1 & 75.6 & 83.4 & 79.1 & 79.3 & 71.5 & 84.8 \\ 
        \rowcolor{green!5}
        VisRAG\_SlideVQA & 85.1 & 92.6 & 54.7 & 84.1 & 87.6 & 87.1 & 91.5 & 92.3 & \textbf{92.6} & 82.8 & 92.1 \\ 
        \rowcolor{green!5}
        VisRAG\_InfoVQA & 83.5 & 88.6 & 66.0 & 82.3 & 87.9 & 84.4 & 89.2 & 90.0 & 89.6 & \textbf{91.8} & 90.8 \\ 
        \rowcolor{green!5}
        VisRAG\_PlotQA & \textbf{79.3} & 76.5 & 62.7 & 75.9 & 69.4 & 68.0 & 72.7 & 73.0 & 72.4 & 68.5 & 58.9 \\ 
        \rowcolor{green!5}
        ViDoSeek-page & 38.1 & 32.6 & 16.3 & 29.1 & 22.5 & 21.2 & 21.3 & 82.0 & \textbf{84.4} & 32.2 & 74.8 \\ 
        \rowcolor{green!5}
        ViDoSeek-doc & 87.5 & \textbf{90.3} & 69.4 & 79.0 & 73.8 & 75.9 & 75.3 & 82.6 & 82.4 & 78.5 & 88.9 \\ 
        \rowcolor{green!5}
        MMLongBench-page & 27.1 & 36.9 & 0.4 & 15.8 & 13.3 & 11.9 & 12.3 & 47.7 & \textbf{51.0} & 32.4 & 47.5 \\ 
        \rowcolor{green!5}
        MMLongBench-doc & 80.4 & \textbf{85.2} & 28.8 & 63.0 & 42.6 & 39.7 & 41.3 & 50.3 & 50.7 & 49.9 & 55.2 \\ 
        \bottomrule
    \end{tabular}
    \end{adjustbox}
\label{tab:app_detailed_score}
\end{table*}

\begin{table*}[htbp]
\centering
\caption{Results on the MMEB-V1 benchmark, which consists of 36 image embedding tasks. IND and OOD denote in-distribution and out-of-distribution datasets, respectively. Some results are adopted from Embed-RL~\cite{embedrl}.}

\renewcommand{\arraystretch}{1.0}
\resizebox{0.8\textwidth}{!}{
\begin{tabular}{lccccccc}
\toprule
\multirow{2}{*}{Model} & \multicolumn{4}{c}{Per Meta-Task Score} & \multicolumn{3}{c}{Average Score} \\ \cmidrule(r){2-5} \cmidrule(l){6-8}
& Classification & VQA & Retrieval & Grounding & IND & OOD & Overall \\ \midrule
\# of Datasets & 10 & 10 & 12 & 4 & 20 & 16 & 36 \\ \midrule
\rowcolor{dt!50}
\multicolumn{8}{c}{\textit{Baseline Models}} \\ \midrule
CLIP~\cite{radford2021learningtransferablevisualmodels} & 42.8 & 9.1 & 53.0 & 51.8 & 37.1 & 38.7 & 37.8 \\
BLIP-2~\cite{li2023blip2bootstrappinglanguageimagepretraining} & 27.0 & 4.2 & 33.9 & 47.0 & 25.3 & 25.1 & 25.2 \\
SigLIP~\cite{zhai2023sigmoidlosslanguageimage} & 40.3 & 8.4 & 31.6 & 59.5 & 32.3 & 38.0 & 34.8 \\
OpenCLIP~\cite{Cherti_2023} & 47.8 & 10.9 & 52.3 & 53.3 & 39.3 & 40.2 & 39.7 \\
UniIR (BLIP$_{\text{FF}}$)~\cite{uniir} & 42.1 & 15.0 & 60.1 & 62.2 & 44.7 & 40.4 & 42.8 \\
UniIR (CLIP$_{\text{SF}}$)~\cite{uniir} & 44.3 & 16.2 & 61.8 & 65.3 & 47.1 & 41.7 & 44.7 \\
Magiclens~\cite{zhang2024magiclensselfsupervisedimageretrieval} & 38.8 & 8.3 & 35.4 & 26.0 & 31.0 & 23.7 & 27.8 \\ \midrule
\rowcolor{dt!50}
\multicolumn{8}{c}{\textit{MLLM-based Baseline Models}} \\ \midrule
VLM2Vec-2B~\cite{vlm2vec} & 59.0 & 49.4 & 65.4 & 73.4 & 66.0 & 52.6 & 60.1 \\
VLM2Vec-7B~\cite{vlm2vec} & 62.6 & 57.8 & 69.9 & 81.7 & 72.2 & 57.8 & 65.8 \\
VLM2Vec-V2~\cite{mmebv2} & 62.9 & 56.3 & 69.5 & 77.3 & 68.8 & 59.9 & 64.9 \\
MMRet-7B~\cite{zhou2024megapairsmassivedatasynthesis} & 56.0 & 57.4 & 69.9 & 83.6 & 68.0 & 59.1 & 64.1 \\
CAFe-V1-7B~\cite{yu2025cafeunifyingrepresentationgeneration} & 65.2 & 65.6 & 70.0 & 91.2 & 75.8 & 62.4 & 69.8 \\
CAFe-V2-7B~\cite{yu2025cafeunifyingrepresentationgeneration} & 63.6 & 61.7 & 69.1 & 87.6 & 72.8 & 61.1 & 67.6 \\
mmE5-11B~\cite{chen2025mme5improvingmultimodalmultilingual} & 67.6 & 62.8 & 70.9 & 89.7 & 72.3 & 66.7 & 69.8 \\
LLaVE-2B~\cite{lan2026llavelargelanguagevision} & 62.1 & 60.2 & 65.2 & 84.9 & 69.4 & 59.8 & 65.2 \\
LLaVE-7B~\cite{lan2026llavelargelanguagevision} & 65.7 & 65.4 & 70.9 & 91.9 & 75.0 & 64.4 & 70.3 \\
UniME-4B~\cite{gu2025breakingmodalitybarrieruniversal} & 54.8 & 55.9 & 64.5 & 81.8 & 68.2 & 52.7 & 64.2 \\
UniME-7B~\cite{gu2025breakingmodalitybarrieruniversal} & 66.8 & 66.6 & 70.6 & 90.9 & 74.6 & 65.8 & 70.7 \\
UME-R1-2B~\cite{ume} & 64.8 & 62.8 & 67.6 & 77.2 & 71.5 & 60.4 & 66.6 \\
UME-R1-7B~\cite{ume} & 67.1 & 69.2 & 71.9 & 84.9 & \textbf{76.1} & 65.1 & 71.3 \\

Embed-RL-2B~\cite{embedrl} & 62.8 & 67.9 & 68.6 & 90.4 & 71.9 & 65.9 & 69.2 \\
Embed-RL-4B~\cite{embedrl} & 63.7 & 70.5 & 71.3 & 91.4 & 74.3 & 67.3 & 71.2 \\
\midrule
\rowcolor{dt!50}
\multicolumn{8}{c}{\textit{Ours}} \\ \midrule

\modelname (Qwen2-VL-2B) & 64.5 & 69.0 & 70.0 & 88.9 & 71.4 & 68.4 & 70.0 \\
\modelname (Qwen3-VL-4B)& \textbf{67.7} & \textbf{74.2} & 
\textbf{74.5} & \textbf{94.9} & 75.4 & \textbf{74.0} & \textbf{74.8} \\

\bottomrule
\end{tabular}
}
\label{tab:app_v1_results}

\end{table*}




\end{document}